\pdfoutput=1

\documentclass[11pt]{article}

\usepackage{ACL2023}

\usepackage{times}
\usepackage{latexsym}

\usepackage[T1]{fontenc}

\usepackage[utf8]{inputenc}

\usepackage{microtype}

\usepackage{inconsolata}

\usepackage{graphicx}
\usepackage{CJKutf8}
\newcommand{\data}{\textsc{VideoQuestions}}
\newcommand{\questions}{2265}
\newcommand{\chapterTitles}{2789}
\usepackage{multirow}

\usepackage{mathabx}
\usepackage{subcaption}
\usepackage{booktabs}

\definecolor[named]{JungleGreen}{cmyk}{0.75, 0, 0.19, 0.33}
\title{ECIS-VQG: Generation of Entity-centric Information-seeking Questions from Videos}

\author{Arpan Phukan$^1$, Manish Gupta$^2$, Asif Ekbal$^3$\\
  $^1$IIT Patna, $^2$Microsoft, $^3$IIT Jodhpur \\
  \texttt{arpan\_2121cs33@iitp.ac.in, gmanish@microsoft.com, asif@iitj.ac.in}
  \\}
  
\begin{document}
\maketitle
\begin{abstract}
Previous studies on question generation from videos have mostly focused on generating questions about common objects and attributes and hence are not entity-centric. In this work, we focus on the generation of entity-centric information-seeking questions from videos. Such a system could be useful for video-based learning, recommending ``People Also Ask'' questions, video-based chatbots, and fact-checking. 
Our work addresses three key challenges: identifying question-worthy information, linking it to entities, and effectively utilizing multimodal signals. Further, to the best of our knowledge, there does not exist a large-scale dataset for this task. Most video question generation datasets are on TV shows, movies, or human activities or lack entity-centric information-seeking questions. Hence, we contribute a diverse dataset of YouTube videos, \data, consisting of 411 videos with \questions{} manually annotated questions. 
We further propose a model architecture combining Transformers, rich context signals (titles, transcripts, captions, embeddings), and a combination of cross-entropy and contrastive loss function to encourage entity-centric question generation.
Our best method yields BLEU, ROUGE, CIDEr, and METEOR scores of 71.3, 78.6, 7.31, and 81.9, respectively, 
demonstrating practical usability.
We make the code and dataset publicly available\footnote{\url{https://github.com/thePhukan/ECIS-VQG/}}.
\end{abstract}


\section{Introduction}
\label{Introduction}

Question Generation (QG) aims at generating a valid and fluent question according to a given passage and an optional target answer. 
QG systems can be used to create interactive learning materials, exams, quizzes~\cite{huang2014tedquiz,krishna2015automatic}, interview questions, clarification questions~\cite{kumar2020clarq} and study aids for students~\cite{wang2018qg,chen2018learningq}. In customer support, QG systems can be helpful in generation of frequently asked questions (FAQs) pages which can help train better chatbots, thereby improving customer support services. Lastly, generating thought-provoking questions can stimulate critical thinking and can play a vital role in formulating research questions, surveys, and interview protocols.

\begin{figure}[!t]
    \centering
    \includegraphics[width=\columnwidth]{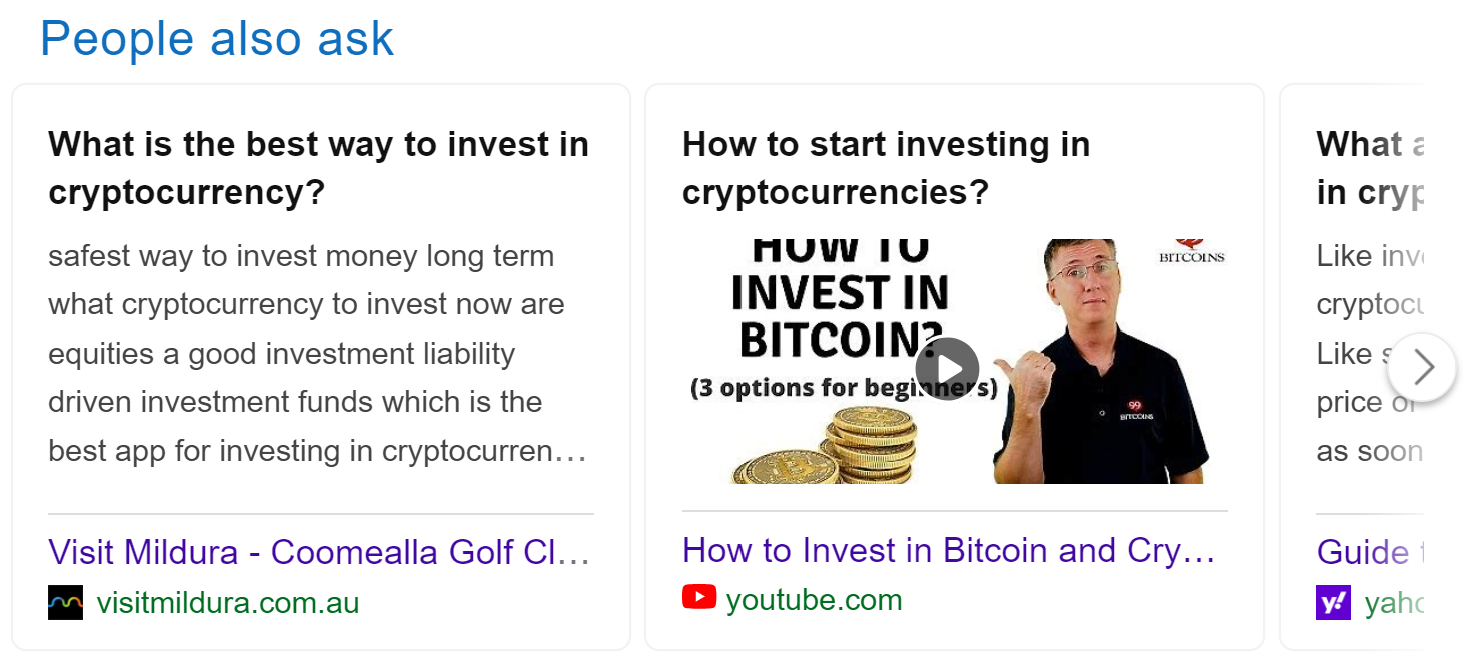}
    \caption{
    Bing's People Also Ask (PAA) module (accessed Sep 21, 2024) displays a question (second one) along with a relevant video thumbnail. When user clicks on the thumbnail, they land on the most relevant chapter within the video. PAA is an apt application for Entity-centric Information-seeking Video QG systems.}
    \label{fig:paa}
\end{figure}

Most of the previous studies in QG are based on text passages~\cite{pan2019recent,zhang2021review}. In contrast, QG is rather underexplored in images and videos. In video QG~\cite{yang2021just,wang2020video,su2021end,guo2020multi}, the goal is to generate meaningful questions about a video, optionally targeting an answer. Although under-explored, video QG has multiple applications in video-based e-learning systems, recommending questions for ``People Also Ask'' module on search engines, video-based chatbots, and fact-checking from videos. Fig.~\ref{fig:paa} shows a video clip (chapter) being shown as an answer to the second question in the ``People Also Ask'' module.

There are a very few studies on video-based QG, even those studies either generate questions only from transcripts~\cite{krishna2015automatic,huang2014tedquiz,priya2022automatic} (and hence do not use visual information from videos) or generate questions about common objects (in a frame of a video) and attributes (such as colour of clothing) present in the video~\cite{yang2021just,wang2020video,guo2020multi,su2021end,gupta2022newskvqa}
. They focus on generating questions like 
``What did the person get a plastic bowl from?'', 
``What will I show you?'',
``What is the nationality of the celebrities highlighted in the last one third of the video?'',
etc. Such research studies involving questions whose answers are too contextually dependent on the video content, do not provide generally applicable information-seeking questions. Note that these questions do not contain entity mentions. 

\begin{figure}[!t]
    \centering
    \includegraphics[width=\columnwidth]{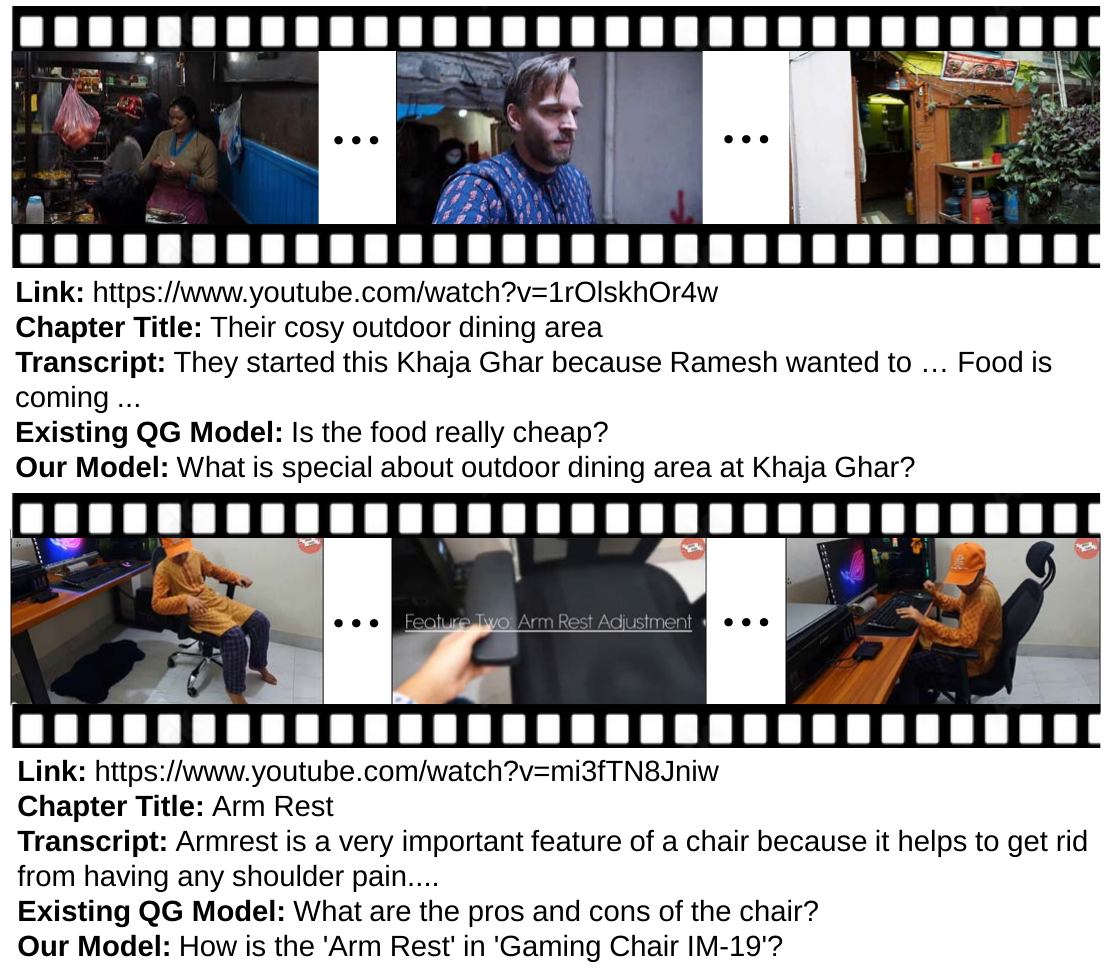}
    \caption{Two examples of ECIS QG task. For example-1, although the existing QG model~\cite{t5_neg_qs} generates a grammatically sound question, it lacks key context information like a place (\textit{Where is the food cheap?}) or subject (\textit{Which food item?}). In example-2, without the particular chair's name, the question generated by the existing QG model is too broad.}
    \label{fig:examples}
\end{figure}

Unlike such studies, we focus on a novel video QG problem setting: generation of entity-centric information-seeking (ECIS) questions from videos. We define an entity as something with an actual existence or a distinct identity. It could refer to objects of focus in a video, encompassing places, buildings, persons, things, or organizations—for example, a book, the company Microsoft, and the president of the USA. 
Our problem definition can be established as follows: Let $V$ denotes a video context consisting of a sequence of frames $\{F_1, F_2, ..., F_n\}$, where each frame $F_i$ represents a snapshot of the video at time $t_i$. Each frame contains visual and possibly textual information.
Given a video context $V$, the goal is to generate a set of questions $Q = \{q_1, q_2, ..., q_m\}$ that specifically targets entities within that context. 
These entities are selected based on their potential to pique a user's interest, and the generated questions aim to extract detailed information about these entities from the video content. Each question $q_i$ is formulated to elicit information about a particular entity $e_j$ and can be represented as a function $Q(e_j)$ that maps the entity $e_j$ to a natural language question.
As illustrated in Fig.~\ref{fig:examples}, in the first example, the traditional QG model (T5 fine-tuned on SQuAD~\cite{t5_neg_qs}) generated the question ``\textit{Is the food really cheap?}''. This question is not self-complete and needs more context like specific ``place'' or the subject (food item). On the other hand, the question generated by our proposed system, ``\textit{What is special about outdoor dining area at Khaja Ghar?}'' is self-complete and information-seeking. The corresponding video provides a relevant answer to this question.

One of the core challenges of generating entity-centric questions lies in effectively extracting suitable features from diverse video content, encompassing a wide array of topics, contexts, and presentation styles. Solely relying upon pre-determined patterns or templates or identifying objects in a video, etc. ~\cite{gupta2022newskvqa}, is not enough. (1) Templates can only lead to bland questions like ``How to make the \_\_\_ dish?'' Generating more interesting questions like ``How to add the Holy Trinity (chopped onion, bell pepper, and celery) in Seafood Gumbo?'' needs accurate understanding and parsing of the video content. (2) Relying only on templates is inefficient because videos from different domains will require custom-tailored templates. Conventional question generation models may struggle to align with the intricate details embedded in the video's visual and textual elements. Video content often spans chapters, with transitions, visual cues, and dynamic pacing, adding complexity to the extraction process. 

In this study, we aim to exploit video chapter titles to generate ECIS questions for videos. However, not all chapter titles are question-worthy. Hence, we train classifiers to find question-worthy chapter titles. Also, chapter titles cannot be directly used as questions; they need to be modified appropriately. Hence, we also fine-tune Transformer~\cite{vaswani2017attention}-based encoder-decoder models like T5~\cite{raffel2020exploring} and BART~\cite{lewis2020bart} on our manually annotated dataset. To generate high-quality ECIS questions, besides chapter title, we supply additional input based on richer context like video title, video transcript, and visual information like frame captions and frame embeddings. Our models are trained using a contrastive loss that motivates the generation of ECIS questions rather than questions about common objects or attributes. Lastly, we also explore effectiveness of prompt engineering with Alpaca LoRA~\cite{taori2023stanford}, Qwen-VL~\cite{bai2023qwen},  GPT-3.5-Turbo~\cite{gpt3.5turbo} and GPT-4o~\cite{gpt4o} for this ECIS-VQG task.

Overall, we make the following contributions.
    (1) We propose a novel problem setting for video QG: entity-centric information-seeking questions.
    (2) We contribute a novel dataset, \data{}\footnote{\url{https://www1.iitp.ac.in/~ai-nlp-ml/resources.html}}, featuring videos across various domains and manually annotated ECIS questions.
    (3) We analyze the efficacy of various Transformer encoder-decoder techniques, prompt engineering methods, multimodal video information encoding approaches, and cross-entropy combined with contrastive loss.
    (4) We observe that BART, fine-tuned using cross-entropy and contrastive loss, performs the best with a multimodal input including chapter and video titles, frame captions, video transcript, and CLIP-based video embeddings.


\section{Related Work}
\subsection{Text-based Question Generation}
Literature reveals that the existing works on QG are mainly focused on text ~\cite{pan2019recent,zhang2021review,mitra2021zero,mitra2020transformer,chatterjee2020faqaugmenter}, and has been studied either at document ~\cite{pan2020semantic,yang2017semi,tuan2020capturing}, paragraph ~\cite{du2018harvesting,zhang2020dual}, sentence level~\cite{ali2010automatic} or keyword level~\cite{pan2020learning}.  

\subsection{Visual Question Generation}

\citet{mostafazadeh2016generating} introduced visual QG~\cite{patil2020visual} to generate questions from an image. Visual questions can be categorized in three groups: (1) visually grounded questions~\cite{antol2015vqa,krishna2017visual}, i.e., questions that can be answered based on information present in the image itself. (2) Commonsense-based questions~\cite{wang2017fvqa,wang2017explicit}, i.e.,  questions that can be answered using a combination of external commonsense knowledge source along with the grounded information in the image. (3) World knowledge-based questions~\cite{shah2019kvqa,penamakuri2023answer}, i.e., questions that can be answered using a combination of external factual knowledge base along with the grounded information in the image. Approaches used for visual QG include encoder-decoder models~\cite{mostafazadeh2016generating,zhang2017automatic}, compositional approaches~\cite{liu2018ivqa,patro2018multimodal,zhang2017automatic}, generative models~\cite{jain2017creativity}, reinforcement learning approaches~\cite{yang2018visual,fan2018reinforcement}, and bilinear pooling models~\cite{fukui2016multimodal,ben2017mutan,li2018visual}. Visual QG has also been studied in domain-specific ways~\cite{mehta2024circuitvqa}.


\subsection{Video Question Generation}

Video QG ~\cite{yang2021just,su2021end} is a challenging task due to the inherent
temporal structure of the video information. These studies generate questions solely from transcripts ~\cite{priya2022automatic} or about common objects and attributes in the video ~\cite{gupta2022newskvqa}. However, generation of ECIS questions (our focus) from videos introduces further challenges: (1) 
ECIS questions should be self-complete rather than generic information about objects and attributes, necessitating the identification of valuable content from videos, which is difficult.
(2) Linking the visual information with an entity and generating a text-based question focused on the entity that could be useful to users is complex. (3) Designing methods that can leverage different kinds of multimodal signals effectively is challenging.

Existing datasets on video QG~\cite{rajpurkar2016squad,gupta2022newskvqa,acharya2019tallyqa} primarily focus on domains like TV shows, films, and human activities while ignoring videos with a knowledge emphasis. Given that we propose a novel ECIS question generation problem setting, there does not exist a large scale dataset for this task. Hence, we first curate a diverse dataset of videos from YouTube, \data{}, consisting of 411 videos with \questions{} manually annotated questions. Videos span a broad spectrum of domains like tech, food, travel, engineering, etc. We developed several models to study the effectiveness of the proposed dataset. Detailed related work is covered in Appendix~\ref{app:relatedWork}.

\section{\data{} Dataset} 
\label{dataset}
\noindent\textbf{Data Curation and Pre-processing:} \label{Data Curation and Preprocessing}
We curate a dataset of videos from YouTube with manually annotated questions as follows. We first gather meta-data for 50K\footnote{
We select 50,000 high-click videos from our internal dataset of videos and corresponding click data.} YouTube videos using YouTube-DL\footnote{\url{https://github.com/ytdl-org/youtube-dl/}}. We only retain videos less than 10 minutes in duration to manage the overall dataset size. \textit{Also, we retain a video only if it has an English transcript}. Our goal is to select videos that are highly information-centric and relevant to real-world topics. Hence, we choose videos belonging to these categories: Education, Entertainment, Howto \& Style, News \& Politics, People \& Blogs, Science \& Technology and Travel \& Events. For each video, we also extract the chapter titles with start and end timestamps, time-stamped transcripts, and video titles from the meta-data. 
Overall, the proposed \data{} dataset contains 411 videos with an average length of $\sim$6 minutes. Category distribution of the videos is as follows: Education: 121, Entertainment: 32, Howto \& Style: 90, News \& Politics: 8, People \& Blogs: 75, Science \& Technology: 65, Travel \& Events: 20. Further, the dataset has overall \chapterTitles{} chapter titles with an average duration of $\sim$48 seconds per chapter.


\begin{figure*}[!t]
    \centering
    \includegraphics[width=\textwidth]{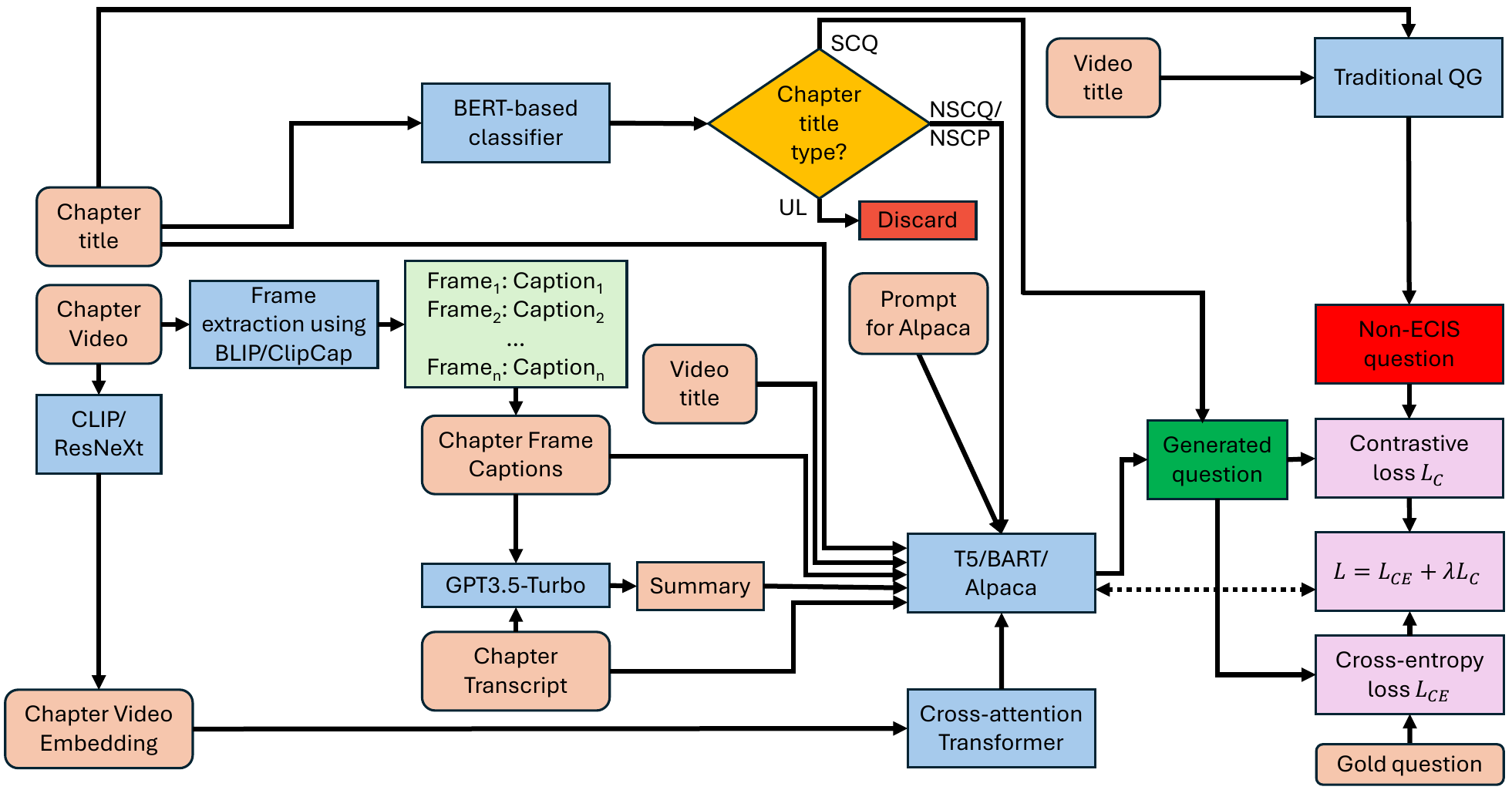}
    
    \caption{Architecture of the proposed method indicating various components like input representations, chapter titles classifier, and Transformer encoder-decoder model. Here, inputs are shown in orange, outputs are in green, models are in blue, and loss functions are in pink. Note that loss computation happens at train time only. Prompt is used for Alpaca only. Cross-attention Transformer layer and video embedding is not used for Alpaca.}
    \label{model}
\end{figure*}

\noindent\textbf{Categorization of Chapter Titles:}
One way to generate ECIS questions from a video is to exploit video chapter titles (demarcated by the uploaders themselves). However, not all chapter titles are question-worthy. Hence, we train classifiers to find question-worthy chapter titles. According to the richness of the information in chapter titles, we categorize them into the following four types.
(1) Useless (UL): Chapter titles that provide no meaningful information or are not conducive to generating coherent and meaningful questions. Examples: ``\textit{Intro}'', ``\textit{Like-share-subscribe.}'' 
(2) Self-Complete Questions (SCQ): Chapter titles that, by themselves, form proper questions related to the specific topics being discussed in the respective video chapter. Examples: ``\textit{What is a Wormhole?}'' or ``\textit{What is Servant Leadership in Business?}''. 
(3) Not Self-Complete Questions (NSCQ): Chapter titles that are appropriate questions but require additional information or context to make sense. Examples: ``\textit{When was she born?}'', ``\textit{What makes each place different?}''. In the context of the respective videos, their complete question equivalents would be ``When was the LA model Elizabeth Turner born?'', ``How is Lakewood Ranch different from Palmer Ranch?'' 
(4) Not Self-Complete Phrases (NSCP):  Chapter titles that contain key information but cannot be used directly as questions. Examples: ``\textit{Gives you Online Visibility}'' or ``\textit{Aerial roots}.'' 

We asked two male Indian annotators in their late 20s to manually classify the \chapterTitles{} chapter titles in \data{} using the above categorization scheme. Both annotators have a Bachelors in Computer Science and Engineering, and have a good grasp of the English language with prior knowledge in the similar tasks. The annotation process resulted in a substantial kappa score of 0.65, indicating a significant level of agreement between our annotators. Conflicts were resolved by mutual discussions between the annotators. Overall, chapter title type distribution is as follows: 524 UL, 133 NSCQ, 189 SCQ and 1943 NSCP. 
Further, for NSCQ and NSCP chapter titles, the annotators were also asked to write out corresponding questions ensuring that the questions were grammatically sound, information-seeking, entity-centric and highly relevant to the respective video chapter. Ignoring the UL type chapter titles, overall the \data{} dataset contains questions corresponding to each of the 133 (NSCQ) + 189 (SCQ) + 1943 (NSCP) = \questions{} chapters.

Specifically, the annotators were provided these guidelines to generate questions.
    (1) Questions should be broadly related to the existing chapter title and not unrelated.
    (2) Ensure that questions are sensible, capable of generating interest in users, are information-seeking, centered around some real world entity, and are directly related to the content in the corresponding video chapter.
    (3) Avoid questions that require additional context or refer to topics beyond the specific chapter. 
    (4) Avoid questions with relative time references like ``current US president'' or ``current Google CEO'' or ``conversion rate USD/JPY'', etc.


\noindent\textbf{Obtaining Frame Captions:}
While transcripts and video titles offer insight into the linguistic content of the video, they cannot capture the rich visual context that videos inherently possess. Hence, we obtain captions for individual video frames. Using frame captions along with other inputs like chapter title, video title, transcript for question generation can ensure that the generated questions retain information from the spoken as well as visual context of the video. 
We obtain frame captions as follows. We first obtain the clip corresponding to a chapter based on its start and end timestamp. Next, we sample one frame per second from the chapter clip. Each extracted frame is then passed through the BLIP \cite{li2022blip}/ClipCap \cite{mokady2021clipcap} model, to generate frame captions. All frame captions corresponding to the same chapter are concatenated. Overall, in our dataset, average lengths of chapter title, video title, transcript and frame captions are around 4.6, 10.3, 132.4 and 173.6 words, respectively.

\section{ECIS Question Generation from Videos}
\label{methodology}

Given a video, at inference time, our proposed system generates ECIS questions from the video as follows. We first extract chapter titles from the video description. Next, using a chapter title classifier (discussed next in this section), we classify each chapter as UL, SCQ, NSCQ, or NSCP. UL chapter titles are discarded. SCQ chapter titles directly serve as relevant ECIS questions. NSCQ and NSCP questions must be processed by a ECIS Questions Generator module (discussed later in this section) to generate appropriate questions. 
We illustrate the proposed method in Fig.~\ref{model}.


\subsection{Chapter Title Classifier} \label{Chapter Title Classifier}
Chapter titles are a good summary of the content in a video chapter. For example, chapter titles, such as ``\textit{Setting up Sendgrid}'' or ``\textit{Meaning Of Wholesaling}'' indicate the information discussed in that specific chapter of the video. On the contrary, titles like ``\textit{intro}'', ``\textit{outro}'', etc., highlight that these will be filler segments and not useful to generate ECIS questions.  
Hence, we train a classifier to classify chapter titles as UL, SCQ, NSCQ, or NSCP. To this end, we build a BERT~\cite{devlin2019bert}-based classifier that takes in the chapter title as input and classifies them into one of the four classes.

We discard UL chapter titles because there is no valuable information in these chapter titles.  These chapters do not hold enough information-rich content, whether in transcripts, titles, or video. Intro (00:00 - 00.36) and Outro (6:25 - 7:04) chapters in \url{https://www.youtube.com/watch?v=y1doijbvWWg}. Similarly, Intro of the video (00:00 - 00:12) and End of the video (2:34 - 2:45) chapters in \url{https://www.youtube.com/watch?v=1rjEVtf6oss}.

\subsection{ECIS Questions Generator} \label{ECIS Questions Generator}

\begin{figure}[!t]
    \centering
    \includegraphics[width=\columnwidth]{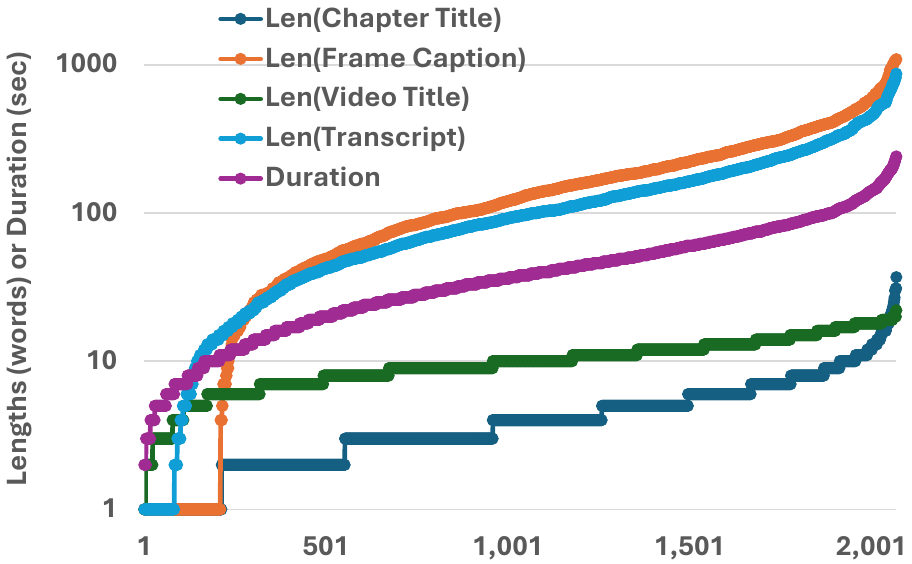}
    \caption{Length distribution (in words) of chapter title, frame captions, video title, transcript; and duration (in seconds) for NSC (NSCQ+NSCP) chapters in the \data{} dataset}
    \label{fig:stats}
\end{figure}

The ECIS question generator converts NSC chapter titles (NSCP+NSCQ) into ECIS questions. We first create a dataset using only NSCP and NSCQ chapter titles. Fig.~\ref{fig:stats} shows the length distribution (in words) and duration (in seconds) for these chapters, all following a power law distribution. The dataset, consisting of 2,076 samples, is then split into 80\% for training, 10\% for validation, and 10\% for testing.

\noindent\textbf{Input Representation:}
Given a video clip corresponding to a chapter, we have several representations and related context: chapter title, video title, chapter transcript, and frame captions. We experiment with various combinations of these signals to form an input for each sample.

Frame captions and transcripts could be long and noisy. We show examples of noisy frame captions and transcripts in Table~\ref{tab:noisy_vt_fc} in Appendix~\ref{gpt3.5Example}. For refining the noisy frame captions and video transcripts, we create summaries using GPT-3.5-Turbo with the following prompt: \textit{``Use the given video transcript and frame captions to describe what is likely happening in the video. Do not add any new keywords that are not present in the given information: [Chapter] Transcript: [Transcript] Frame captions: [Caption]''}. We provide an example of generated summary in Appendix~\ref{gpt3.5Example}. Lastly, to better capture the rich visual semantics in the video, we encode the chapter clip using CLIP~\cite{radford2021learning} or ResNeXt~\cite{xie2017aggregated} models. 

\noindent\textbf{Transformer encoder-decoder methods:}
To effectively integrate visual information from frames with the textual context, we propose a multimodal approach based on Transformer based encoder-decoder models like BART~\cite{lewis2020bart} and T5~\cite{raffel2020exploring}. We fine-tune these models using various combinations of the input features to produce questions. 
As part of input combinations, when we use video embeddings, we do the following. For ResNeXt, we employ a linear layer to reduce the size of the embeddings from 2048 dimensions to 1024. In the case of CLIP embeddings, we use a linear layer to expand the dimensions from 512 to 1024. Next, we combine the text input embeddings with the video embedding using a multi-head cross-attention Transformer layer where the transformed video embedding forms the key and value, while text tokens form the query. By incorporating text tokens into the query, we give higher weight to the video aspects related to the text input. It helps us downplay the irrelevant parts of the video and amplify those aspects related to the text input.
The video-influenced text tokens are then combined with positional embeddings and fed as input to the first BART/T5 encoder layer. In the subsequent encoder layers, we combine the hidden representation of the previous encoder layer with the video embedding using the same multi-head cross-attention Transformer layer where the transformed video embedding forms the key and value, while embeddings of the previous encoder layer tokens form the query.
The last BART/T5 encoder feeds into the decoder layers. The last BART/T5 decoder layer generates the output one token at a time in an auto-regressive manner.
Our experiments show that performing such fusion across all encoder layers was effective for BART. But for T5, performing such fusion only at the last encoder layer led to better performance.



\begin{table*}[!t]
\scriptsize
\centering
\begin{tabular}{|p{0.75in}|p{0.5in}|p{1in}|p{1.5in}|p{1in}|p{1.25in}|}
\hline
VideoID&Chapter&Chapter title&Video title& non-ECIS Question&ECIS Generated Question\\ \hline
\hline
1rjEVtf6oss&[120, 154]&Branch cutting with roots&How to Take Cuttings of Arrowhead Plant& What will we do with the roots of the plant? &Do you have to cut the branch from the root of Syngonium Cuttings?\\ \hline
Du\_gNCQ5QYY&[134,216]& Two groups of members & Our Mighty Network Adventure Begins - The Inside Story of Building a Thriving Online Community & What is the name of the new online community? & What kind of groups does Creative Life Center support?
\\ \hline
\end{tabular}
\caption{
Examples of non-ECIS questions and those generated by our ``ECIS questions generator'' for two chapters.} 
\label{negative_sample_qs}
\end{table*}

\noindent\textbf{Fine-tuning using combination of Contrastive and Cross-entropy Loss:}\label{Fine-tuning using combination of Contrastive and Cross-entropy Loss}
We fine-tune the encoder-decoder model using the typical cross-entropy loss. However, we also want to enhance the distinctiveness between our ECIS questions and the generic non-ECIS questions about common objects. To achieve this, we design a contrastive loss which attempts to generate questions, different compared to a non-ECIS question. Thus, our model is fine-tuned to minimize the overall loss given by a combination of the cross-entropy loss $\mathcal{L}_{CE}$ and contrastive loss $\mathcal{L}_C$: $\mathcal{L}=\mathcal{L}_{CE}+\lambda\mathcal{L}_C$ where $\lambda$ balances the two loss components. 
Contrastive loss is computed as follows. For a sample chapter, we generate non-ECIS question using a model~\cite{t5_neg_qs} fine-tuned for traditional QG which hopefully generates questions with common objects and attributes. The input to this model is chapter title and video title. Contrastive loss then tries to ensure that a question generated by our ECIS questions generator is at least margin $m$ away from this non-ECIS question. Note that the traditional non-ECIS QG generator is frozen and not fine-tuned along with the ECIS question generator. 
Sometimes, traditional question generation model could lead to ECIS questions rather than non-ECIS ones. In such cases, it would be wrong to ensure that the question generated by the ECIS QG generator is different from this question. We handle such situations as follows. 
Typically, ECIS questions contain entity names, while non-ECIS questions contain common nouns rather than entity names. Hence, we generate question candidates for the chapter titles and use a named entity recognition (NER) check to pick the candidates which are more likely to be non-ECIS. This is achieved by avoiding candidates which have even one token labeled as an entity with confidence greater than a threshold $c$ as measured by the BERT-based NER model, ``\textit{dslim/bert-base-NER}'' ~\cite{sang2003introduction}
Such a method ensures that we use non-ECIS questions along with questions generated by our ``ECIS questions generator'' model as part of contrastive loss computation.
Table \ref{negative_sample_qs} shows examples of non-ECIS questions and question generated by our ``ECIS questions generator'' for two chapter titles. It shows that non-ECIS questions are too generic and refer to common names like ``plant'' and ``online community'' while generated ECIS questions are self-complete with appropriate entity names.



\begin{table*}[!t]
\scriptsize
\centering
\begin{tabular}{|l|l|c|c|c|c|c|c|c|}
\hline
&Model&BLEU-1&CIDEr&METEOR&Distinct-1&Distinct-2&BERT-Score&ROUGE-L\\ \hline\hline
\multirow{6}{*}{A}&\citet{yang2021just}&4.5&0.805&23.2&35.9&78.1&54.2&15.9\\ \cline{2-9}
&\citet{lopez2020transformer}&4.6&0.767&21.0&37.2&80.7&53.3&15.5\\ \cline{2-9}
&\citet{vqg_baseline_lmqg}&6.4&0.779&28.1&46.3&84.9&59.3&21.8\\ \cline{2-9}
&\citet{t5_neg_qs}&8.2&0.923&24.4&35.4&75.0&52.6&20.2\\ \cline{2-9}
&Llama3-8B P1\cite{llama3modelcard}&0.8&0.162&7.9&51.0&83.3&35.76&5.3\\ \cline{2-9}
&Llama3-8B P2\cite{llama3modelcard}&1.5&0.220&11.9&50.48&85.5&40.6&8.1\\ \hline
\hline
\multirow{3}{*}{B}&Alpaca-p2~\cite{taori2023stanford}&22.4&0.844&31.9&38.9&80.0&58.4&22.0\\
\cline{2-9}
&Qwen-VL~\cite{bai2023qwen}&2.7&0.260&31.8&25.9&70.4&56.8&17.2\\
\cline{2-9}
&GPT-3.5-Turbo-p2~\cite{gpt3.5turbo}&7.5&1.053&35.9&35.7&80.6&62.9&24.6\\ \cline{2-9}
&GPT-4o~\cite{gpt4o}&7.1&0.870&41.6&29.1&77.0&64.7&25.6\\
\cline{2-9}
&Llama3-8B P2\cite{llama3modelcard}&19.8&2.334&45.5&45.0&82.8&68.0&38.1\\ \hline
\hline
\multirow{4}{*}{C}&$T5_{C}$(C, V, F, T)&26.1&2.782&51.3&48.2&87.7&73.8&43.6\\ \cline{2-9}
&$T5_{C}$(C, V, S(F, T))&29.4&3.218&61.9&49.0&88.5&80.3&46.5\\ \cline{2-9}
&$T5_{CC}$(C, V, F, T)&59.4&5.999&72.8&47.9&87.9&85.1&69.5\\ \cline{2-9}
&$T5_{CC}$(C, V, S(F, T))&49.3&5.511&70.7&49.1&88.5&85.1&62.2\\ \hline
\hline
\multirow{4}{*}{D}&$B_{C}$(C, V, F, T)&29.7&3.152&54.4&47.7&86.8&75.3&48.0\\ \cline{2-9}
&$B_{C}$(C, V, S(F, T))&28.2&3.639&56.4&49.7&88.6&73.5&43.2\\ \cline{2-9}
&$B_{CC}$(C, V, F, T)&54.4&5.746&71.0&48.6&87.5&84.4&66.9\\ \cline{2-9}
&$B_{CC}$(C, V, S(F, T))&67.8&7.125&\textbf{83.5}&\textbf{50.4}&\textbf{88.9}&\textbf{91.3}&76.9\\ \hline\hline
\multirow{1}{*}{\textbf{E}}&$B_{CC}$(C, V, S(F, T), $E_C$)&\textbf{71.3}&\textbf{7.311}&81.9&47.2&87.6&90.0&\textbf{78.6}\\ \hline
\end{tabular}
\caption{ECIS question generation results. 
B=BART-large, T5=T5-base, $X_{C}$=Model X was trained using Cross-Entropy Loss, $X_{CC}$=Model X was trained using Contrastive and Cross-Entropy Loss. C=Chapter Title, V=Video Title, F=Frame Caption, T=Transcript and S(F,T)=Summary of F and T, generated by GPT-3.5-Turbo. $E_C$=CLIP-based embeddings. RL values for $B_{CC}$(C, V, S(F, T), $E_C$) are statistically significant over $B_{CC}$(C, V, S(F, T)) (two-tailed t-test, p<0.05) signifying that video signals indeed help with ECIS QG.}
\label{tab:ECISQGresults}
\end{table*}

\noindent\textbf{Prompt engineering:}
We experiment with Alpaca~\cite{taori2023stanford} and GPT-3.5-Turbo~\cite{gpt3.5turbo} by designing standard prompt template while providing all combinations of four features: transcript, frame captions, chapter title and video title. For example, considering chapter title and transcript as input, prompt could be: ``For the paragraph given below, bot is provided with the following attributes from a video: Chapter Title: [Chapter Title] and Transcript: [Transcript]. The question is then generated 
by utilizing the video data provided. The generated question should not contain any kind of pronouns. The generated question's answer must be in the Chapter Title or Transcript. 
These should be engaging, information-seeking, lengthy and centered around some real world entity.''
We discuss rationale and design choices 
in the Appendix~\ref{promt_engineering}. Further, we also tried combinations of inputs with prompts where we provided positive or negative exemplar questions. Prompt with positive exemplars augmented the above base prompt with this text: ``THE GENERATED QUESTIONS MUST BE LIKE: `What is the history of the world’s largest hybrid tensegrity, Kurilpa Bridge?', `Does Syngonium Plant grow easily from properly taken cutting?', `How can you eat at a Khaja Ghar in
Nepal?', `How to wrap rice in banana leaf?', `What is the estimated net worth of Elizabeth Turner?', `Is it possible to travel through a wormhole, if they exist?' '' Prompt with negative exemplars augmented the above base prompt with this text: ``DO NOT GENERATE QUESTIONS LIKE: `What is the name of this place?', `Where do I need to take cuttings?', `What does she look like?', `What does dad do?', `What is her current net worth?', `What would happen if you take a trip to one of these wormholes?' '' More details about prompt engineering are in the Appendix~\ref{promt_engineering}. We also experiment with Qwen-VL~\cite{bai2023qwen} and GPT-4o~\cite{gpt4o} models.
\section{Experiments and Results}
\subsection{Chapter Titles Classifier Results}
For the chapter titles classifier, we use the ``bert-base-uncased'' model. The last six layers of BERT provide input to an MLP (Multi-Layered Perceptron), with 4608 input nodes, with hidden layers consisting of 1536 and 768 nodes, respectively. For regularization and non-linearity, we apply dropout (0.2) and ReLU activation for the hidden layers.
To address class imbalance, we incorporate class weights using the formula $\frac{N}{C_{n}\times X_{i}}$, where $N$ represents the total number of samples, $C_{n}$ is the number of classes (4 in our case), and $X_{i}$ is the number of instances for a particular class. By incorporating $X_{i}$ in the denominator, we ensure that more frequent classes receive less weight, effectively mitigating the impact of class imbalance.
For fine-tuning our classifier, we employ the Cross-Entropy loss, AdamW optimizer and train the model for 100 epochs. Following the training phase, our classifier yielded an accuracy of 0.934. Table \ref{classification_report} (in the Appendix) presents the class-wise precision, recall, and F1 scores.



\subsection{ECIS Questions Generation Results} \label{ECIS Questions Generation Results}

For the ECIS video QG task, we conduct comprehensive experiments utilizing various input feature combinations, namely chapter title, chapter transcript, frame captions, and video title. We aim to assess the quality of generated outputs for different models, including BART, T5, and Alpaca, under different input variations. To evaluate the performance, we calculate several metrics, including BERT-Score~\cite{zhang2019bertscore}, CIDEr~\cite{vedantam2015cider}, METEOR~\cite{banerjee2005meteor}, BLEU~\cite{papineni2002bleu}, ROUGE~\cite{lin2004rouge}, and Distinct~\cite{li2016diversity} scores.




For training, we employ batch sizes of 4 and 2 as per memory constraints of different input combinations. We set the maximum input tokens to 2048. 
We train our models for 50 epochs using an A100-40GB NVIDIA graphics card. We set $\lambda$=1 and NER threshold $c$=0.4. When using frame captions, across several experiments, we found that BLIP-based captions led to better results compared to ClipCap; hence, we report all the results using BLIP.
Table~\ref{tab:ECISQGresults} shows the main results for the ECIS question generation task. Results are shown in five blocks: Block A is for baselines, Block B is for prompt engineering methods, Block C is for variations of T5, Block D is for variations of BART and Block E shows improvements using video embeddings. We discuss ablations later in this section. Note that we show results for BART-Large and T5-Base, both of which have 12 layers in the encoder as well as in the decoder.

Comparing Block A with C and D, we observe that our proposed method with T5 and BART outperforms existing baseline methods by massive margins. Note that though these baselines are old, unfortunately, there does not exist better (or recent) baselines for the video question generation problem. Further, comparing Block B with C and D, we observe that fine-tuning Transformer-based encoder-decoder models are better than prompt engineering with Alpaca and GPT-3.5-Turbo. Note that we experiment with 19 different prompts for Alpaca and GPT-3.5-Turbo and report the best results (using prompt p2) in the table. Detailed prompt engineering results are in Appendix~\ref{promt_engineering}. We also experiment with Qwen-VL~\cite{bai2023qwen} and GPT-4o~\cite{gpt4o} models. It is evident that our best model outperforms these zero-shot state-of-the-art models. To illustrate this, we highlight a few samples in Table~\ref{tab:comparisonQwenGPT4o} in the Appendix for a clear comparison.

Additionally, to ensure fairness, we fine-tuned baseline models using our training data. Table~\ref{tab:baseline-finetuning} in the Appendix illustrates that our best model outperforms these fine-tuned baseline models as well.

Comparing blocks C and D indicates that BART typically leads to better results compared to T5. We observe that a combination of contrastive loss and cross-entropy loss is better than using cross-entropy alone. Also, within the BART block, we observe that using a summary of frame captions and transcripts using GPT-3.5-Turbo is better than using the noisy  original versions.

Lastly, comparing Block E with Block D, we observe that additional video embedding input (CLIP-based embedding) leads to improvements for BART, indicating the importance of effective encoding of the visual information in video clips. Overall, the best model (Block E) leads to a BLEU-1 of 71.3, CIDEr of 7.311, METEOR of 81.9, BERT-Score of 90.0 and a ROUGE-L score of 78.6 underlining its practical usability.

\begin{table}[!t]
\scriptsize
\centering
\begin{tabular}{|l|p{0.4in}|p{0.5in}|l|}
\hline
Model & Context Relevance & Engagement Index & Fluency \\ \hline\hline
Alpaca-p2 & 2.52 & 2.82 & 3.44 \\ \hline

GPT-3.5-Turbo-p2 & 3.69 & 3.72 & \textbf{3.99} \\ \hline

$T5_{CC}$(C, V, F, T) & 3.68 & 3.71 & 3.61 \\ \hline

$T5_{CC}$(C, V, S(F, T)) & 3.69 & 3.72 & 3.62 \\ \hline

$B_{CC}$(C, V, F, T) & 3.69 & 3.74 & 3.62 \\ \hline

$B_{CC}$(C, V, S(F, T)) & 3.75 & 3.84 & 3.72 \\ \hline

$B_{CC}$(C, V, S(F, T), $E_C$) & \textbf{3.91} & \textbf{3.97} & \textbf{3.99} \\ \hline

\end{tabular}
\caption{Human evaluation results
}
\label{human_evaluation}
\end{table}

\noindent\textbf{Ablation for Input Representations}: Table~\ref{tab:ablations} in Appendix~\ref{app:ablation} shows results for ECIS QG using different input combinations. We also show results using BART/T5 and using just cross-entropy loss vs a combination of contrastive and cross-entropy loss. Introducing contrastive loss yields superior results compared to exclusively relying on cross-entropy. This trend is observed across all the models with different input combinations.
We also observe that models with the input features (C,V,T) and (C,V,S(F,T)) achieve the best results, with (C,V,T) performing exceptionally well in BLEU-1, CIDEr and ROUGE-L. Although (C,V,S(F,T)) is slightly inferior in automatic evaluation metrics, it generates higher-quality questions according to the human evaluation results as shown in Table~\ref{human_evaluation}.
Further, for our best model, $B_{CC}$(C, V, S(F, T)) as well as the equivalent T5 variant, we tried two variations by including video embeddings obtained using CLIP or ResNeXt as an additional input.
Table~\ref{tab:ablation2} in Appendix~\ref{app:ablation} indicates that CLIP-based video embeddings outperform ResNeXt-based ones for BART, while ResNeXt embeddings are better for T5. Overall, $B_{CC}(C, V, S(F, T))$ with $E_C$ yields the optimal performance.

\subsection{Qualitative Analysis}









\noindent\textbf{Human Evaluation}:
Table~\ref{human_evaluation} shows human evaluation scores for prompt engineering methods (Alpaca and GPT-3.5-Turbo) and also for the best T5 and BART models. It is evident that $B_{CC}$(C, V, S(F, T),$E_C$) demonstrates better performance for generating task-conforming questions. Detailed evaluation protocol and results for more methods are shown in Table~\ref{tab:human_evaluation_all} in the appendix.

\noindent\textbf{Case Studies}: 
Tables~\ref{tab:good_examples} and~\ref{tab:good_examples_APPENDIX} in Appendix~\ref{app:qualitative} showcase examples of questions generated by our best method. 
The tables demonstrate the model's adeptness at integrating information from various sources—including video titles, chapter titles, and visual data—to produce relevant, engaging, and fluent questions that capture viewer interest.

\noindent\textbf{Error Analysis}: 
To understand the generation errors of our model, we manually categorized 100 generated questions based on the kind of errors, if any. We found four kinds of errors: (1) HA: Questions with Hallucinations (irrelevant/unrelated key words) (2) GR: grammatically incorrect questions (3) MK: Ambiguous questions or questions missing imperative keywords to make it an ECIS question (4) CX: 
Contextually incongruous questions which are not exactly related to the topic discussed in the video chapter. Tables~\ref{bad_examples} and~\ref{bad_examples_APPENDIX} in Appendix~\ref{app:qualitative} show a few of such examples. Amongst the overall errors, we found the following error type distribution: HA (38\%), GR (11\%), MK (27\%) and CX (24\%).


\section{Conclusion}
In this paper, we have proposed an interesting problem for generating ECIS questions from videos. Since there was no readily available dataset, we create a new dataset, \data{}, with 411 videos and more than 2K manually annotated questions. We proposed a system which consumes rich text and visual signals to arrive at an effective video representation. This, in turn, helps train a Transformer encoder-decoder model with cross entropy and contrastive loss. Extensive experiments show that our best model generates high quality ECIS questions which can be leveraged for several practical applications like ``Video-based People Also Ask'', ``Online tutoring systems'', etc. 
In the future, we aim to expand this system to locate relevant clips within videos and broadening our approach to support multiple languages to increase the model's global relevance.




\section{Limitations}
While our research has made significant strides in entity-centric information-seeking question generation from videos, some limitations warrant acknowledgment.

\noindent\textbf{Monolingual:} Our experiments were conducted exclusively in English, thereby restricting the generalizability of our findings to other languages. Given the multilingual nature of online content consumption, it is imperative to scale our methodologies to accommodate a broader linguistic spectrum. Expanding our approach to encompass multiple languages would enhance our model's applicability and cater to a more diverse user base.

\noindent\textbf{Dataset Generalization:} While carefully curated and annotated, the dataset used in our experiments may not fully capture the diversity and complexity present in real-world video content. Future research could utilize more extensive and diverse datasets, encompassing a broader range of topics, styles, and sources. Such datasets would enable a more robust evaluation of our model's performance across different video genres and content types.

\noindent\textbf{Limited dataset size:} The dataset used in this work is relatively small, with only 2265 manually annotated questions, which may limit the model's generalizability. Further research with more extensive and more diverse datasets is needed.

\noindent\textbf{Challenges in handling complex questions:}
 The model may struggle to generate complex questions that require reasoning or inference over multiple attributes or entities. Further research is needed to develop models that can handle these questions.

\section{Ethics Statement}

This research adheres to strict ethical data collection, handling, and dissemination principles. All procedures involving human subjects, including the acquisition of annotations and user-generated content, were conducted with utmost respect for their rights and dignity. Informed consent was obtained from all participants, who were fully informed about the research's objectives and voluntarily chose to participate. This research has also undergone assessment and received approval from our Institutional Review Board (IRB).

Furthermore, our commitment to transparency and reproducibility is paramount. We have meticulously documented all methodologies, data pre-processing steps, model architectures, and hyperparameters, making them publicly available. This ensures accountability within our research and facilitates the replication and validation of our findings by other scholars, thereby fostering trust within the scientific community.

In acknowledging the potential biases within the dataset and subsequent models, we have taken several proactive measures to mitigate them. Firstly, we have included video samples from various geographical locations (USA, Nepal, South Korea, Japan, etc.), ensuring diversity in the dataset. Additionally, the video data we have utilized represents a range of cultural backgrounds, further increasing inclusivity. Moreover, we have incorporated a linguistically diverse video corpus into our analysis (English, Japanese, Korean, Arabic, etc.). These steps, coupled with thorough evaluations, have been taken to actively minimize the generation of discriminatory or offensive outputs by our model.

Moreover, to safeguard the ethical use of the dataset, access will be granted only upon completion of an agreement stipulating that the data will be utilized solely for research purposes. Additionally, human experts involved in data annotation and evaluation are compensated by institute policies, ensuring fair treatment and acknowledgment of their contributions.

\bibliography{custom}

\begin{thebibliography}{92}
\expandafter\ifx\csname natexlab\endcsname\relax\def\natexlab#1{#1}\fi

\bibitem[{Acharya et~al.(2019)Acharya, Kafle, and Kanan}]{acharya2019tallyqa}
Manoj Acharya, Kushal Kafle, and Christopher Kanan. 2019.
\newblock Tallyqa: Answering complex counting questions.
\newblock In \emph{Proceedings of the AAAI conference on artificial intelligence}, volume~33, pages 8076--8084.

\bibitem[{AI@Meta(2024)}]{llama3modelcard}
AI@Meta. 2024.
\newblock \href {https://github.com/meta-llama/llama3/blob/main/MODEL_CARD.md} {Llama 3 model card}.

\bibitem[{Ali et~al.(2010)Ali, Chali, and Hasan}]{ali2010automatic}
Husam Ali, Yllias Chali, and Sadid~A Hasan. 2010.
\newblock Automatic question generation from sentences.
\newblock In \emph{Actes de la 17e conf{\'e}rence sur le Traitement Automatique des Langues Naturelles. Articles courts}, pages 213--218.

\bibitem[{Antol et~al.(2015)Antol, Agrawal, Lu, Mitchell, Batra, Zitnick, and Parikh}]{antol2015vqa}
Stanislaw Antol, Aishwarya Agrawal, Jiasen Lu, Margaret Mitchell, Dhruv Batra, C~Lawrence Zitnick, and Devi Parikh. 2015.
\newblock Vqa: Visual question answering.
\newblock In \emph{Proceedings of the IEEE international conference on computer vision}, pages 2425--2433.

\bibitem[{Bai et~al.(2023)Bai, Bai, Yang, Wang, Tan, Wang, Lin, Zhou, and Zhou}]{bai2023qwen}
Jinze Bai, Shuai Bai, Shusheng Yang, Shijie Wang, Sinan Tan, Peng Wang, Junyang Lin, Chang Zhou, and Jingren Zhou. 2023.
\newblock Qwen-vl: A versatile vision-language model for understanding, localization, text reading, and beyond.
\newblock \emph{arXiv preprint arXiv:2308.12966}.

\bibitem[{Bajaj et~al.(2016)Bajaj, Campos, Craswell, Deng, Gao, Liu, Majumder, McNamara, Mitra, Nguyen et~al.}]{bajaj2016ms}
Payal Bajaj, Daniel Campos, Nick Craswell, Li~Deng, Jianfeng Gao, Xiaodong Liu, Rangan Majumder, Andrew McNamara, Bhaskar Mitra, Tri Nguyen, et~al. 2016.
\newblock Ms marco: A human generated machine reading comprehension dataset.
\newblock \emph{arXiv preprint arXiv:1611.09268}.

\bibitem[{Banerjee and Lavie(2005)}]{banerjee2005meteor}
Satanjeev Banerjee and Alon Lavie. 2005.
\newblock Meteor: An automatic metric for mt evaluation with improved correlation with human judgments.
\newblock In \emph{Proceedings of the acl workshop on intrinsic and extrinsic evaluation measures for machine translation and/or summarization}, pages 65--72.

\bibitem[{Bao et~al.(2018)Bao, Gong, Duan, Zhou, and Zhao}]{bao2018question}
Junwei Bao, Yeyun Gong, Nan Duan, Ming Zhou, and Tiejun Zhao. 2018.
\newblock Question generation with doubly adversarial nets.
\newblock \emph{IEEE/ACM Transactions on Audio, Speech, and Language Processing}, 26(11):2230--2239.

\bibitem[{Ben-Younes et~al.(2017)Ben-Younes, Cadene, Cord, and Thome}]{ben2017mutan}
Hedi Ben-Younes, R{\'e}mi Cadene, Matthieu Cord, and Nicolas Thome. 2017.
\newblock Mutan: Multimodal tucker fusion for visual question answering.
\newblock In \emph{Proceedings of the IEEE international conference on computer vision}, pages 2612--2620.

\bibitem[{Chai and Wan(2020)}]{chai2020learning}
Zi~Chai and Xiaojun Wan. 2020.
\newblock Learning to ask more: Semi-autoregressive sequential question generation under dual-graph interaction.
\newblock In \emph{Proceedings of the 58th Annual Meeting of the Association for Computational Linguistics}, pages 225--237.

\bibitem[{Chatterjee et~al.(2020)Chatterjee, Gupta, and Agrawal}]{chatterjee2020faqaugmenter}
Ankush Chatterjee, Manish Gupta, and Puneet Agrawal. 2020.
\newblock Faqaugmenter: suggesting questions for enterprise faq pages.
\newblock In \emph{Proceedings of the 13th international conference on web search and data mining}, pages 829--832.

\bibitem[{Chen et~al.(2018)Chen, Yang, Hauff, and Houben}]{chen2018learningq}
Guanliang Chen, Jie Yang, Claudia Hauff, and Geert-Jan Houben. 2018.
\newblock Learningq: a large-scale dataset for educational question generation.
\newblock In \emph{Proceedings of the International AAAI Conference on Web and Social Media}, volume~12.

\bibitem[{Chen et~al.(2020)Chen, Wu, and Zaki}]{chen2020reinforcement}
Yu~Chen, Lingfei Wu, and Mohammed~J Zaki. 2020.
\newblock Reinforcement learning based graph-to-sequence model for natural question generation.
\newblock In \emph{International Conference on Learning Representations}.

\bibitem[{Chen et~al.(2023)Chen, Wu, and Zaki}]{chen2023toward}
Yu~Chen, Lingfei Wu, and Mohammed~J Zaki. 2023.
\newblock Toward subgraph-guided knowledge graph question generation with graph neural networks.
\newblock \emph{IEEE Transactions on Neural Networks and Learning Systems}.

\bibitem[{Cheng et~al.(2021)Cheng, Li, Liu, Zhao, Li, Lin, and Zheng}]{cheng2021guiding}
Yi~Cheng, Siyao Li, Bang Liu, Ruihui Zhao, Sujian Li, Chenghua Lin, and Yefeng Zheng. 2021.
\newblock Guiding the growth: Difficulty-controllable question generation through step-by-step rewriting.
\newblock In \emph{Proceedings of the 59th Annual Meeting of the Association for Computational Linguistics and the 11th International Joint Conference on Natural Language Processing (Volume 1: Long Papers)}, pages 5968--5978.

\bibitem[{Devlin et~al.(2019)Devlin, Chang, Lee, and Toutanova}]{devlin2019bert}
Jacob Devlin, Ming-Wei Chang, Kenton Lee, and Kristina Toutanova. 2019.
\newblock Bert: Pre-training of deep bidirectional transformers for language understanding.
\newblock In \emph{Proceedings of naacL-HLT}, volume~1, page~2.

\bibitem[{Dong et~al.(2019)Dong, Yang, Wang, Wei, Liu, Wang, Gao, Zhou, and Hon}]{dong2019unified}
Li~Dong, Nan Yang, Wenhui Wang, Furu Wei, Xiaodong Liu, Yu~Wang, Jianfeng Gao, Ming Zhou, and Hsiao-Wuen Hon. 2019.
\newblock Unified language model pre-training for natural language understanding and generation.
\newblock \emph{Advances in neural information processing systems}, 32.

\bibitem[{Du and Cardie(2018)}]{du2018harvesting}
Xinya Du and Claire Cardie. 2018.
\newblock Harvesting paragraph-level question-answer pairs from wikipedia.
\newblock In \emph{Proceedings of the 56th Annual Meeting of the Association for Computational Linguistics (Volume 1: Long Papers)}, pages 1907--1917.

\bibitem[{Du et~al.(2017)Du, Shao, and Cardie}]{du2017learning}
Xinya Du, Junru Shao, and Claire Cardie. 2017.
\newblock Learning to ask: Neural question generation for reading comprehension.
\newblock In \emph{Proceedings of the 55th Annual Meeting of the Association for Computational Linguistics (Volume 1: Long Papers)}. Association for Computational Linguistics.

\bibitem[{Duan et~al.(2017)Duan, Tang, Chen, and Zhou}]{duan2017question}
Nan Duan, Duyu Tang, Peng Chen, and Ming Zhou. 2017.
\newblock Question generation for question answering.
\newblock In \emph{Proceedings of the 2017 conference on empirical methods in natural language processing}, pages 866--874.

\bibitem[{Fan et~al.(2018)Fan, Wei, Wang, Liu, and Huang}]{fan2018reinforcement}
Zhihao Fan, Zhongyu Wei, Siyuan Wang, Yang Liu, and Xuan-Jing Huang. 2018.
\newblock A reinforcement learning framework for natural question generation using bi-discriminators.
\newblock In \emph{Proceedings of the 27th International Conference on Computational Linguistics}, pages 1763--1774.

\bibitem[{Fukui et~al.(2016)Fukui, Park, Yang, Rohrbach, Darrell, and Rohrbach}]{fukui2016multimodal}
Akira Fukui, Dong~Huk Park, Daylen Yang, Anna Rohrbach, Trevor Darrell, and Marcus Rohrbach. 2016.
\newblock Multimodal compact bilinear pooling for visual question answering and visual grounding.
\newblock In \emph{Proceedings of the 2016 Conference on Empirical Methods in Natural Language Processing}. Association for Computational Linguistics.

\bibitem[{Guo et~al.(2020)Guo, Zhao, Jin, Wei, Yang, Wang, and Yuan}]{guo2020multi}
Zhaoyu Guo, Zhou Zhao, Weike Jin, Zhicheng Wei, Min Yang, Nannan Wang, and Nicholas~Jing Yuan. 2020.
\newblock Multi-turn video question generation via reinforced multi-choice attention network.
\newblock \emph{IEEE Transactions on Circuits and Systems for Video Technology}, 31(5):1697--1710.

\bibitem[{Gupta and Gupta(2022)}]{gupta2022newskvqa}
Pranay Gupta and Manish Gupta. 2022.
\newblock Newskvqa: Knowledge-aware news video question answering.
\newblock In \emph{Pacific-asia conference on knowledge discovery and data mining}, pages 3--15. Springer.

\bibitem[{Heilman and Smith(2010)}]{heilman2010good}
Michael Heilman and Noah~A Smith. 2010.
\newblock Good question! statistical ranking for question generation.
\newblock In \emph{Human language technologies: The 2010 annual conference of the North American Chapter of the Association for Computational Linguistics}, pages 609--617.

\bibitem[{Hosking and Riedel(2019)}]{hosking2019evaluating}
Tom Hosking and Sebastian Riedel. 2019.
\newblock Evaluating rewards for question generation models.
\newblock In \emph{Proceedings of the 2019 Conference of the North American Chapter of the Association for Computational Linguistics: Human Language Technologies, Volume 1 (Long and Short Papers)}, pages 2278--2283.

\bibitem[{Huang et~al.(2014)Huang, Tseng, Sun, and Chen}]{huang2014tedquiz}
Yi-Ting Huang, Ya-Min Tseng, Yeali~S Sun, and Meng~Chang Chen. 2014.
\newblock Tedquiz: automatic quiz generation for ted talks video clips to assess listening comprehension.
\newblock In \emph{2014 IEEE 14Th international conference on advanced learning technologies}, pages 350--354. IEEE.

\bibitem[{Jain et~al.(2017)Jain, Zhang, and Schwing}]{jain2017creativity}
Unnat Jain, Ziyu Zhang, and Alexander~G Schwing. 2017.
\newblock Creativity: Generating diverse questions using variational autoencoders.
\newblock In \emph{Proceedings of the IEEE conference on computer vision and pattern recognition}, pages 6485--6494.

\bibitem[{Krishna et~al.(2015)Krishna, Bhowmick, Ghosh, Sahu, and Roy}]{krishna2015automatic}
Amrith Krishna, Plaban Bhowmick, Krishnendu Ghosh, Archana Sahu, and Subhayan Roy. 2015.
\newblock Automatic generation and insertion of assessment items in online video courses.
\newblock In \emph{Proceedings of the 20th International Conference on Intelligent User Interfaces Companion}, pages 1--4.

\bibitem[{Krishna et~al.(2017)Krishna, Zhu, Groth, Johnson, Hata, Kravitz, Chen, Kalantidis, Li, Shamma et~al.}]{krishna2017visual}
Ranjay Krishna, Yuke Zhu, Oliver Groth, Justin Johnson, Kenji Hata, Joshua Kravitz, Stephanie Chen, Yannis Kalantidis, Li-Jia Li, David~A Shamma, et~al. 2017.
\newblock Visual genome: Connecting language and vision using crowdsourced dense image annotations.
\newblock \emph{International journal of computer vision}, 123:32--73.

\bibitem[{Kumar and Black(2020)}]{kumar2020clarq}
Vaibhav Kumar and Alan~W Black. 2020.
\newblock Clarq: A large-scale and diverse dataset for clarification question generation.
\newblock In \emph{Proceedings of the 58th Annual Meeting of the Association for Computational Linguistics}, pages 7296--7301.

\bibitem[{Kumar et~al.(2019)Kumar, Joshi, Mukherjee, Ramakrishnan, and Jyothi}]{kumar2019cross}
Vishwajeet Kumar, Nitish Joshi, Arijit Mukherjee, Ganesh Ramakrishnan, and Preethi Jyothi. 2019.
\newblock Cross-lingual training for automatic question generation.
\newblock In \emph{Proceedings of the 57th Annual Meeting of the Association for Computational Linguistics}, pages 4863--4872.

\bibitem[{Lee et~al.(2020)Lee, Lee, Jeong, Kim, and Hwang}]{lee2020generating}
Dong~Bok Lee, Seanie Lee, Woo~Tae Jeong, Donghwan Kim, and Sung~Ju Hwang. 2020.
\newblock Generating diverse and consistent qa pairs from contexts with information-maximizing hierarchical conditional vaes.
\newblock In \emph{Proceedings of the 58th Annual Meeting of the Association for Computational Linguistics}, pages 208--224.

\bibitem[{Lewis et~al.(2020)Lewis, Liu, Goyal, Ghazvininejad, Mohamed, Levy, Stoyanov, and Zettlemoyer}]{lewis2020bart}
Mike Lewis, Yinhan Liu, Naman Goyal, Marjan Ghazvininejad, Abdelrahman Mohamed, Omer Levy, Veselin Stoyanov, and Luke Zettlemoyer. 2020.
\newblock Bart: Denoising sequence-to-sequence pre-training for natural language generation, translation, and comprehension.
\newblock In \emph{Proceedings of the 58th Annual Meeting of the Association for Computational Linguistics}, pages 7871--7880.

\bibitem[{Li et~al.(2016)Li, Galley, Brockett, Gao, and Dolan}]{li2016diversity}
Jiwei Li, Michel Galley, Chris Brockett, Jianfeng Gao, and William~B Dolan. 2016.
\newblock A diversity-promoting objective function for neural conversation models.
\newblock In \emph{Proceedings of the 2016 Conference of the North American Chapter of the Association for Computational Linguistics: Human Language Technologies}, pages 110--119.

\bibitem[{Li et~al.(2022)Li, Li, Xiong, and Hoi}]{li2022blip}
Junnan Li, Dongxu Li, Caiming Xiong, and Steven Hoi. 2022.
\newblock Blip: Bootstrapping language-image pre-training for unified vision-language understanding and generation.
\newblock In \emph{International Conference on Machine Learning}, pages 12888--12900. PMLR.

\bibitem[{Li et~al.(2018)Li, Duan, Zhou, Chu, Ouyang, Wang, and Zhou}]{li2018visual}
Yikang Li, Nan Duan, Bolei Zhou, Xiao Chu, Wanli Ouyang, Xiaogang Wang, and Ming Zhou. 2018.
\newblock Visual question generation as dual task of visual question answering.
\newblock In \emph{Proceedings of the IEEE conference on computer vision and pattern recognition}, pages 6116--6124.

\bibitem[{Lin(2004)}]{lin2004rouge}
Chin-Yew Lin. 2004.
\newblock Rouge: A package for automatic evaluation of summaries.
\newblock In \emph{Proc. Workshop on Text Summariation Branches Out, Post-Conference Workshop of ACL 2004}, pages 74--81.

\bibitem[{Lindberg et~al.(2013)Lindberg, Popowich, Nesbit, and Winne}]{lindberg2013generating}
David Lindberg, Fred Popowich, John Nesbit, and Phil Winne. 2013.
\newblock Generating natural language questions to support learning on-line.
\newblock In \emph{Proceedings of the 14th European Workshop on Natural Language Generation}, pages 105--114.

\bibitem[{Liu et~al.(2019)Liu, Zhao, Niu, Lai, He, Wei, and Xu}]{liu2019learning}
Bang Liu, Mingjun Zhao, Di~Niu, Kunfeng Lai, Yancheng He, Haojie Wei, and Yu~Xu. 2019.
\newblock Learning to generate questions by learningwhat not to generate.
\newblock In \emph{The world wide web conference}, pages 1106--1118.

\bibitem[{Liu et~al.(2018)Liu, Xiang, Hospedales, Yang, and Sun}]{liu2018ivqa}
Feng Liu, Tao Xiang, Timothy~M Hospedales, Wankou Yang, and Changyin Sun. 2018.
\newblock ivqa: Inverse visual question answering.
\newblock In \emph{Proceedings of the IEEE Conference on Computer Vision and Pattern Recognition}, pages 8611--8619.

\bibitem[{Lopez et~al.(2020)Lopez, Cruz, Cruz, and Cheng}]{lopez2020transformer}
Luis~Enrico Lopez, Diane~Kathryn Cruz, Jan Christian~Blaise Cruz, and Charibeth Cheng. 2020.
\newblock Transformer-based end-to-end question generation.
\newblock \emph{arXiv preprint arXiv:2005.01107}, 4.

\bibitem[{Mazidi and Nielsen(2014)}]{mazidi2014linguistic}
Karen Mazidi and Rodney Nielsen. 2014.
\newblock Linguistic considerations in automatic question generation.
\newblock In \emph{Proceedings of the 52nd Annual Meeting of the Association for Computational Linguistics (Volume 2: Short Papers)}, pages 321--326.

\bibitem[{Mehta et~al.(2024)Mehta, Singh, Varma, and Gupta}]{mehta2024circuitvqa}
Rahul Mehta, Bhavyajeet Singh, Vasudeva Varma, and Manish Gupta. 2024.
\newblock Circuitvqa: A visual question answering dataset for electrical circuit images.
\newblock In \emph{Joint European Conference on Machine Learning and Knowledge Discovery in Databases}, pages 440--460. Springer.

\bibitem[{Mitra et~al.(2020)Mitra, Gupta, and Dandapat}]{mitra2020transformer}
Rajarshee Mitra, Manish Gupta, and Sandipan Dandapat. 2020.
\newblock Transformer models for recommending related questions in web search.
\newblock In \emph{Proceedings of the 29th ACM International Conference on Information \& Knowledge Management}, pages 2153--2156.

\bibitem[{Mitra et~al.(2021)Mitra, Jain, Veerubhotla, and Gupta}]{mitra2021zero}
Rajarshee Mitra, Rhea Jain, Aditya~Srikanth Veerubhotla, and Manish Gupta. 2021.
\newblock Zero-shot multi-lingual interrogative question generation for" people also ask" at bing.
\newblock In \emph{Proceedings of the 27th ACM SIGKDD Conference on Knowledge Discovery \& Data Mining}, pages 3414--3422.

\bibitem[{Mokady et~al.(2021)Mokady, Hertz, and Bermano}]{mokady2021clipcap}
Ron Mokady, Amir Hertz, and Amit~H Bermano. 2021.
\newblock {ClipCap}: Clip prefix for image captioning.
\newblock \emph{arXiv preprint arXiv:2111.09734}.

\bibitem[{Mostafazadeh et~al.(2016)Mostafazadeh, Misra, Devlin, Mitchell, He, and Vanderwende}]{mostafazadeh2016generating}
Nasrin Mostafazadeh, Ishan Misra, Jacob Devlin, Margaret Mitchell, Xiaodong He, and Lucy Vanderwende. 2016.
\newblock Generating natural questions about an image.
\newblock In \emph{Proceedings of the 54th Annual Meeting of the Association for Computational Linguistics (Volume 1: Long Papers)}, pages 1802--1813.

\bibitem[{OpenAI(2024)}]{gpt4o}
OpenAI. 2024.
\newblock \href {https://openai.com/index/hello-gpt-4o/} {Hello gpt-4o}.

\bibitem[{Pan et~al.(2019)Pan, Lei, Chua, and Kan}]{pan2019recent}
Liangming Pan, Wenqiang Lei, Tat-Seng Chua, and Min-Yen Kan. 2019.
\newblock Recent advances in neural question generation.
\newblock \emph{arXiv preprint arXiv:1905.08949}.

\bibitem[{Pan et~al.(2020{\natexlab{a}})Pan, Xie, Feng, Chua, and Kan}]{pan2020semantic}
Liangming Pan, Yuxi Xie, Yansong Feng, Tat-Seng Chua, and Min-Yen Kan. 2020{\natexlab{a}}.
\newblock Semantic graphs for generating deep questions.
\newblock In \emph{Proceedings of the 58th Annual Meeting of the Association for Computational Linguistics}, pages 1463--1475.

\bibitem[{Pan et~al.(2020{\natexlab{b}})Pan, Hu, Chen, Xiang, and Wang}]{pan2020learning}
Youcheng Pan, Baotian Hu, Qingcai Chen, Yang Xiang, and Xiaolong Wang. 2020{\natexlab{b}}.
\newblock Learning to generate diverse questions from keywords.
\newblock In \emph{ICASSP 2020-2020 IEEE International Conference on Acoustics, Speech and Signal Processing (ICASSP)}, pages 8224--8228. IEEE.

\bibitem[{Papineni et~al.(2002)Papineni, Roukos, Ward, and Zhu}]{papineni2002bleu}
Kishore Papineni, Salim Roukos, Todd Ward, and Wei-Jing Zhu. 2002.
\newblock Bleu: a method for automatic evaluation of machine translation.
\newblock In \emph{Proceedings of the 40th annual meeting of the Association for Computational Linguistics}, pages 311--318.

\bibitem[{Patil and Patwardhan(2020)}]{patil2020visual}
Charulata Patil and Manasi Patwardhan. 2020.
\newblock Visual question generation: The state of the art.
\newblock \emph{ACM Computing Surveys (CSUR)}, 53(3):1--22.

\bibitem[{Patro et~al.(2018)Patro, Kumar, Kurmi, and Namboodiri}]{patro2018multimodal}
Badri~N Patro, Sandeep Kumar, Vinod~K Kurmi, and Vinay~P Namboodiri. 2018.
\newblock Multimodal differential network for visual question generation.
\newblock In \emph{2018 Conference on Empirical Methods in Natural Language Processing, EMNLP 2018}, pages 4002--4012. Association for Computational Linguistics.

\bibitem[{Penamakuri et~al.(2023)Penamakuri, Gupta, Gupta, and Mishra}]{penamakuri2023answer}
Abhirama~Subramanyam Penamakuri, Manish Gupta, Mithun~Das Gupta, and Anand Mishra. 2023.
\newblock Answer mining from a pool of images: towards retrieval-based visual question answering.
\newblock In \emph{Proceedings of the Thirty-Second International Joint Conference on Artificial Intelligence}, pages 1312--1321.

\bibitem[{Peng et~al.(2023)Peng, Wu, Allard, Kilpatrick, and Heidel}]{gpt3.5turbo}
Andrew Peng, Michael Wu, John Allard, Logan Kilpatrick, and Steven Heidel. 2023.
\newblock \href {https://openai.com/blog/gpt-3-5-turbo-fine-tuning-and-api-updates} {[link]}.

\bibitem[{Priya et~al.(2022)Priya, Priya, Jenneyl, and Uma}]{priya2022automatic}
T~Janani Priya, KP~Sabari Priya, L~Raxxelyn Jenneyl, and KV~Uma. 2022.
\newblock Automatic question generation from video.
\newblock In \emph{International Conference on Computational Intelligence in Pattern Recognition}, pages 366--372. Springer.

\bibitem[{Radford et~al.(2021)Radford, Kim, Hallacy, Ramesh, Goh, Agarwal, Sastry, Askell, Mishkin, Clark et~al.}]{radford2021learning}
Alec Radford, Jong~Wook Kim, Chris Hallacy, Aditya Ramesh, Gabriel Goh, Sandhini Agarwal, Girish Sastry, Amanda Askell, Pamela Mishkin, Jack Clark, et~al. 2021.
\newblock Learning transferable visual models from natural language supervision.
\newblock In \emph{International conference on machine learning}, pages 8748--8763. PMLR.

\bibitem[{Raffel et~al.(2020)Raffel, Shazeer, Roberts, Lee, Narang, Matena, Zhou, Li, and Liu}]{raffel2020exploring}
Colin Raffel, Noam Shazeer, Adam Roberts, Katherine Lee, Sharan Narang, Michael Matena, Yanqi Zhou, Wei Li, and Peter~J Liu. 2020.
\newblock Exploring the limits of transfer learning with a unified text-to-text transformer.
\newblock \emph{The Journal of Machine Learning Research}, 21(1):5485--5551.

\bibitem[{Rajpurkar et~al.(2016)Rajpurkar, Zhang, Lopyrev, and Liang}]{rajpurkar2016squad}
Pranav Rajpurkar, Jian Zhang, Konstantin Lopyrev, and Percy Liang. 2016.
\newblock Squad: 100,000+ questions for machine comprehension of text.
\newblock In \emph{Proceedings of the 2016 Conference on Empirical Methods in Natural Language Processing}, pages 2383--2392.

\bibitem[{Rao and Daum{\'e}~III(2019)}]{rao2019answer}
Sudha Rao and Hal Daum{\'e}~III. 2019.
\newblock Answer-based adversarial training for generating clarification questions.
\newblock In \emph{Proceedings of the 2019 Conference of the North American Chapter of the Association for Computational Linguistics: Human Language Technologies, Volume 1 (Long and Short Papers)}, pages 143--155.

\bibitem[{Reddy et~al.(2019)Reddy, Chen, and Manning}]{reddy2019coqa}
Siva Reddy, Danqi Chen, and Christopher~D Manning. 2019.
\newblock Coqa: A conversational question answering challenge.
\newblock \emph{Transactions of the Association for Computational Linguistics}, 7:249--266.

\bibitem[{Romero(2021)}]{t5_neg_qs}
Manuel Romero. 2021.
\newblock T5 (base) fine-tuned on squad for qg via ap.
\newblock \url{https://huggingface.co/mrm8488/t5-base-finetuned-question-generation-ap}.

\bibitem[{Sang and De~Meulder(2003)}]{sang2003introduction}
Erik Tjong~Kim Sang and Fien De~Meulder. 2003.
\newblock Introduction to the conll-2003 shared task: Language-independent named entity recognition.
\newblock In \emph{Proceedings of the Seventh Conference on Natural Language Learning at HLT-NAACL 2003}, pages 142--147.

\bibitem[{Scialom et~al.(2019)Scialom, Piwowarski, and Staiano}]{scialom2019self}
Thomas Scialom, Benjamin Piwowarski, and Jacopo Staiano. 2019.
\newblock Self-attention architectures for answer-agnostic neural question generation.
\newblock In \emph{Proceedings of the 57th annual meeting of the Association for Computational Linguistics}, pages 6027--6032.

\bibitem[{Shah et~al.(2019)Shah, Mishra, Yadati, and Talukdar}]{shah2019kvqa}
Sanket Shah, Anand Mishra, Naganand Yadati, and Partha~Pratim Talukdar. 2019.
\newblock Kvqa: Knowledge-aware visual question answering.
\newblock In \emph{Proceedings of the AAAI conference on artificial intelligence}, volume~33, pages 8876--8884.

\bibitem[{Song et~al.(2018)Song, Wang, Hamza, Zhang, and Gildea}]{song2018leveraging}
Linfeng Song, Zhiguo Wang, Wael Hamza, Yue Zhang, and Daniel Gildea. 2018.
\newblock Leveraging context information for natural question generation.
\newblock In \emph{Proceedings of the 2018 Conference of the North American Chapter of the Association for Computational Linguistics: Human Language Technologies, Volume 2 (Short Papers)}, pages 569--574.

\bibitem[{Su et~al.(2020)Su, Xu, Dai, Ji, Yu, and Fung}]{su2020multi}
Dan Su, Yan Xu, Wenliang Dai, Ziwei Ji, Tiezheng Yu, and Pascale Fung. 2020.
\newblock Multi-hop question generation with graph convolutional network.
\newblock \emph{Proceedings of Findings of the Association for Computational Linguistics (EMNLP)}.

\bibitem[{Su et~al.(2021)Su, Chang, Shen, Wang, Chang, Chang, Cheng, and Hsu}]{su2021end}
Hung-Ting Su, Chen-Hsi Chang, Po-Wei Shen, Yu-Siang Wang, Ya-Liang Chang, Yu-Cheng Chang, Pu-Jen Cheng, and Winston~H Hsu. 2021.
\newblock End-to-end video question-answer generation with generator-pretester network.
\newblock \emph{IEEE Transactions on Circuits and Systems for Video Technology}, 31(11):4497--4507.

\bibitem[{Taori et~al.(2023)Taori, Gulrajani, Zhang, Dubois, Li, Guestrin, Liang, and Hashimoto}]{taori2023stanford}
Rohan Taori, Ishaan Gulrajani, Tianyi Zhang, Yann Dubois, Xuechen Li, Carlos Guestrin, Percy Liang, and Tatsunori~B. Hashimoto. 2023.
\newblock Stanford alpaca: An instruction-following llama model.
\newblock \url{https://github.com/tatsu-lab/stanford_alpaca}.

\bibitem[{Tuan et~al.(2020)Tuan, Shah, and Barzilay}]{tuan2020capturing}
Luu~Anh Tuan, Darsh Shah, and Regina Barzilay. 2020.
\newblock Capturing greater context for question generation.
\newblock In \emph{Proceedings of the AAAI conference on artificial intelligence}, volume~34, pages 9065--9072.

\bibitem[{Ushio et~al.(2022)Ushio, Alva-Manchego, and Camacho-Collados}]{vqg_baseline_lmqg}
Asahi Ushio, Fernando Alva-Manchego, and Jose Camacho-Collados. 2022.
\newblock \href {https://doi.org/10.18653/v1/2022.emnlp-main.42} {Generative language models for paragraph-level question generation}.
\newblock In \emph{Proceedings of the 2022 Conference on Empirical Methods in Natural Language Processing}, pages 670--688, Abu Dhabi, United Arab Emirates. Association for Computational Linguistics.

\bibitem[{Vaswani et~al.(2017)Vaswani, Shazeer, Parmar, Uszkoreit, Jones, Gomez, Kaiser, and Polosukhin}]{vaswani2017attention}
Ashish Vaswani, Noam Shazeer, Niki Parmar, Jakob Uszkoreit, Llion Jones, Aidan~N Gomez, {\L}ukasz Kaiser, and Illia Polosukhin. 2017.
\newblock Attention is all you need.
\newblock \emph{Advances in neural information processing systems}, 30.

\bibitem[{Vedantam et~al.(2015)Vedantam, Lawrence~Zitnick, and Parikh}]{vedantam2015cider}
Ramakrishna Vedantam, C~Lawrence~Zitnick, and Devi Parikh. 2015.
\newblock Cider: Consensus-based image description evaluation.
\newblock In \emph{Proceedings of the IEEE conference on computer vision and pattern recognition}, pages 4566--4575.

\bibitem[{Wang et~al.(2020{\natexlab{a}})Wang, Wang, Tao, Zhang, and Xu}]{wang2020neural}
Bingning Wang, Xiaochuan Wang, Ting Tao, Qi~Zhang, and Jingfang Xu. 2020{\natexlab{a}}.
\newblock Neural question generation with answer pivot.
\newblock In \emph{Proceedings of the AAAI conference on artificial intelligence}, volume~34, pages 9138--9145.

\bibitem[{Wang et~al.(2021)Wang, Ji, Sun, Yang, and Sakai}]{wang2021mirtt}
Junjie Wang, Yatai Ji, Jiaqi Sun, Yujiu Yang, and Tetsuya Sakai. 2021.
\newblock {MIRTT}: learning multimodal interaction representations from trilinear transformers for visual question answering.
\newblock In \emph{Findings of the Association for Computational Linguistics: EMNLP 2021}, pages 2280--2292.

\bibitem[{Wang et~al.(2017{\natexlab{a}})Wang, Wu, Shen, Dick, and Van Den~Henge}]{wang2017explicit}
Peng Wang, Qi~Wu, Chunhua Shen, Anthony Dick, and Anton Van Den~Henge. 2017{\natexlab{a}}.
\newblock Explicit knowledge-based reasoning for visual question answering.
\newblock In \emph{Proceedings of the 26th International Joint Conference on Artificial Intelligence}, pages 1290--1296.

\bibitem[{Wang et~al.(2017{\natexlab{b}})Wang, Wu, Shen, Dick, and Van Den~Hengel}]{wang2017fvqa}
Peng Wang, Qi~Wu, Chunhua Shen, Anthony Dick, and Anton Van Den~Hengel. 2017{\natexlab{b}}.
\newblock Fvqa: Fact-based visual question answering.
\newblock \emph{IEEE transactions on pattern analysis and machine intelligence}, 40(10):2413--2427.

\bibitem[{Wang et~al.(2019)Wang, Feng, Wang, and Zhang}]{wang2019answer}
Weichao Wang, Shi Feng, Daling Wang, and Yifei Zhang. 2019.
\newblock Answer-guided and semantic coherent question generation in open-domain conversation.
\newblock In \emph{Proceedings of the 2019 Conference on Empirical Methods in Natural Language Processing and the 9th International Joint Conference on Natural Language Processing (EMNLP-IJCNLP)}, pages 5066--5076.

\bibitem[{Wang et~al.(2020{\natexlab{b}})Wang, Su, Chang, Liu, and Hsu}]{wang2020video}
Yu-Siang Wang, Hung-Ting Su, Chen-Hsi Chang, Zhe-Yu Liu, and Winston~H Hsu. 2020{\natexlab{b}}.
\newblock Video question generation via semantic rich cross-modal self-attention networks learning.
\newblock In \emph{ICASSP 2020-2020 IEEE International Conference on Acoustics, Speech and Signal Processing (ICASSP)}, pages 2423--2427. IEEE.

\bibitem[{Wang et~al.(2018)Wang, Lan, Nie, Waters, Grimaldi, and Baraniuk}]{wang2018qg}
Zichao Wang, Andrew~S Lan, Weili Nie, Andrew~E Waters, Phillip~J Grimaldi, and Richard~G Baraniuk. 2018.
\newblock Qg-net: a data-driven question generation model for educational content.
\newblock In \emph{Proceedings of the fifth annual ACM conference on learning at scale}, pages 1--10.

\bibitem[{Xiao et~al.(2021)Xiao, Zhang, Li, Sun, Tian, Wu, and Wang}]{xiao2021ernie}
Dongling Xiao, Han Zhang, Yukun Li, Yu~Sun, Hao Tian, Hua Wu, and Haifeng Wang. 2021.
\newblock Ernie-gen: an enhanced multi-flow pre-training and fine-tuning framework for natural language generation.
\newblock In \emph{Proceedings of the Twenty-Ninth International Conference on International Joint Conferences on Artificial Intelligence}, pages 3997--4003.

\bibitem[{Xie et~al.(2017)Xie, Girshick, Doll{\'a}r, Tu, and He}]{xie2017aggregated}
Saining Xie, Ross Girshick, Piotr Doll{\'a}r, Zhuowen Tu, and Kaiming He. 2017.
\newblock Aggregated residual transformations for deep neural networks.
\newblock In \emph{Proceedings of the IEEE conference on computer vision and pattern recognition}, pages 1492--1500.

\bibitem[{Yang et~al.(2021)Yang, Miech, Sivic, Laptev, and Schmid}]{yang2021just}
Antoine Yang, Antoine Miech, Josef Sivic, Ivan Laptev, and Cordelia Schmid. 2021.
\newblock Just ask: Learning to answer questions from millions of narrated videos.
\newblock In \emph{Proceedings of the IEEE/CVF international conference on computer vision}, pages 1686--1697.

\bibitem[{Yang et~al.(2018)Yang, Lu, Lee, Batra, and Parikh}]{yang2018visual}
Jianwei Yang, Jiasen Lu, Stefan Lee, Dhruv Batra, and Devi Parikh. 2018.
\newblock Visual curiosity: Learning to ask questions to learn visual recognition.
\newblock In \emph{Conference on Robot Learning}, pages 63--80. PMLR.

\bibitem[{Yang et~al.(2017)Yang, Hu, Salakhutdinov, and Cohen}]{yang2017semi}
Zhilin Yang, Junjie Hu, Ruslan Salakhutdinov, and William Cohen. 2017.
\newblock Semi-supervised qa with generative domain-adaptive nets.
\newblock In \emph{Proceedings of the 55th Annual Meeting of the Association for Computational Linguistics (Volume 1: Long Papers)}, pages 1040--1050.

\bibitem[{Yao et~al.(2018)Yao, Zhang, Luo, Tao, and Wu}]{yao2018teaching}
Kaichun Yao, Libo Zhang, Tiejian Luo, Lili Tao, and YanJun Wu. 2018.
\newblock Teaching machines to ask questions.
\newblock In \emph{Proceedings of the 27th International Joint Conference on Artificial Intelligence}, pages 4546--4552.

\bibitem[{Zhang et~al.(2021)Zhang, Guo, Chen, Fan, and Cheng}]{zhang2021review}
Ruqing Zhang, Jiafeng Guo, Lu~Chen, Yixing Fan, and Xueqi Cheng. 2021.
\newblock A review on question generation from natural language text.
\newblock \emph{ACM Transactions on Information Systems (TOIS)}, 40(1):1--43.

\bibitem[{Zhang et~al.(2020)Zhang, Guo, Fan, Lan, and Cheng}]{zhang2020dual}
Ruqing Zhang, Jiafeng Guo, Yixing Fan, Yanyan Lan, and Xueqi Cheng. 2020.
\newblock Dual-factor generation model for conversation.
\newblock \emph{ACM Transactions on Information Systems (TOIS)}, 38(3):1--31.

\bibitem[{Zhang et~al.(2017)Zhang, Qu, You, Yang, and Zhang}]{zhang2017automatic}
Shijie Zhang, Lizhen Qu, Shaodi You, Zhenglu Yang, and Jiawan Zhang. 2017.
\newblock Automatic generation of grounded visual questions.
\newblock In \emph{Proceedings of the 26th International Joint Conference on Artificial Intelligence}, pages 4235--4243.

\bibitem[{Zhang et~al.(2019)Zhang, Kishore, Wu, Weinberger, and Artzi}]{zhang2019bertscore}
Tianyi Zhang, Varsha Kishore, Felix Wu, Kilian~Q Weinberger, and Yoav Artzi. 2019.
\newblock Bertscore: Evaluating text generation with bert.
\newblock \emph{arXiv preprint arXiv:1904.09675}.

\end{thebibliography}
\bibliographystyle{acl_natbib}

\appendix

\section{Annotator Details}

\begin{table}
\scriptsize\centering
\begin{tabular}{|l|l|l|l|}
\hline
 & Precision & Recall & F1 \\ \hline
NSCQ & 0.44 & 0.64 & 0.52 \\ \hline
SCQ & 0.76 & 0.62 & 0.68  \\ \hline
UL & 0.27 & 1.00 & 0.43 \\ \hline
NSCP & 1.00 & 0.98 & 0.99  \\ \hline
 \hline
Accuracy & \multicolumn{3}{r|}{0.93} \\ \hline
Macro Avg & 0.62 & 0.81 & 0.65  \\ \hline
Weighted Avg & 0.95 & 0.93 & 0.94 \\ \hline
\end{tabular}
\caption{Classification results of Bert-based model on test set}\label{classification_report}
\end{table}

For the chapter title classifier annotation, we compensated the annotators at the standard rates used by universities. Two of the annotators are our regular employees who receive a monthly salary at a rate, 
adhering to the university's guidelines. Additionally, we recruited contract-based annotators who were remunerated at per the standard rates. 

\section{Example summary generated by GPT 3.5-Turbo}
\label{gpt3.5Example}
We use GPT-3.5-Turbo to generate summary of frame captions and transcript for a chapter. Here is an example.

\begin{table*}[!t]
\scriptsize
\centering
\begin{tabular}{|p{0.45\textwidth}|p{0.45\textwidth}|}
\hline
\multicolumn{1}{|c|}{\textbf{Chapter Transcripts}} & \multicolumn{1}{c|}{\textbf{Frame Captions}}  \\ \hline
And the shorts...they did happen. 3 days ago, a top post on the Bitcoin subreddit was about BTC shorts being up over 900 percent….. What contributed to this!? Because Chico thinks they are about to GET Rektity, REKT. How so Chico!?…..it is a market manipulators' pattern, which is known throughout all markets, where the controllers...make & 
twitter's twitter account is shown in the screen

the twitter page for the twitter page, showing the twitter page and the twitter page

a man in a gray shirt and black hat is sitting in front of a black wall with a

a man in a gray shirt is sitting in front of a black wall with a blue and white

the twitter screen shows the number of the space shuttles

the twitter screen shows the chart of the star wars

the twitter screen shows the twitter screen showing the twitter screen

...
\\ \hline
you cannot imagine how many bad courses are out there and i decided to create the best course in the world but you might ask yourself why you should buy my course and not choose other courses because i am x google x amazon x facebook and many many other x & google chrome chrome chrome chrome chrome chrome chrome chrome chrome chrome chrome chrome chrome chrome chrome chrome chrome chrome

a blue background with a yellow border

just the best - screenshot

a man sitting at a desk with a laptop

a man in a blue shirt is sitting at a desk
\\ \hline
podcasts, you likely end your writing or recording session with more content than will go into the finished product. Even if you feel like what you’ve edited out is unnecessary, your members might really enjoy the long-form version. This is kind of a funny example, but take Zach Snyder’s cut of Justice League.& 
membership membership membership membership membership membership membership membership membership membership membership membership membership membership membership

membership membership membership membership membership membership membership membership membership membership membership membership membership membership membership membership

a man in a black shirt is standing in front of a purple wall

a man in a black shirt is standing in front of a purple background

...
\\ \hline
Nutrition & 

a blue background with a white and green logo

keto pep seris - peri - calories - calories - calories -

nutrition nutrition nutrition nutrition nutrition nutrition nutrition nutrition nutrition nutrition nutrition nutrition nutrition nutrition nutrition nutrition nutrition nutrition nutrition

a pink background with the words thank for watching please, my channel

thank card with a pink background and white text
\\ \hline
For beginners chocolate decorations For the pattern prepare You can print the free template. Check the description box below. Zipper bag Adhesive tape & 
chocolate fond fond fond fond fond fond fond fond fond fond fond fond fond fond fond fond fond fond

chocolate cutters

for the pattern prepare

a black background with the words for the word, for the people

a person is drawing a pattern on a piece of paper

a person is holding a piece of paper with a drawing of a pattern

...

\\ \hline
by default, ourDB will keep 3 copies of backup for each user & 

a screen shot of a computer with a number of different items

a screen shot of a cell with the text `how to use the phone '

the screen of a cell phone with the text, `how to use the phone? '

the screen shows the settings and settings for the app

the screen of a computer with a green screen
\\ \hline
\end{tabular}
\caption{Noisy Video Transcripts and Frame Captions}\label{tab:noisy_vt_fc}
\end{table*}

\noindent\textbf{Prompt:} Use the given video transcript and frame captions to describe what is likely happening in the video. Do not add any new keywords that are not present in the given information:

\noindent\textbf{Video Transcript:} ``Wormhole, a theoretical gateway to travel distant places in the universe almost instantly. It can even be used to travel to different multiverses, of course in theory... But is it possible or safe for humans to even go nearby? We are speculating about ``Can you travel through a Wormhole?'' in today's video of DiscoverZen. A Wormhole or Einstein–Rosen bridge is a speculative structure linking disparate points in spacetime and is based on a special solution of the Einstein field equations. Wormholes are consistent with the general theory of relativity by Einstein, but they are yet to be discovered practically. Many scientists postulate that wormholes are merely projections of a fourth spatial dimension, analogous to how a two-dimensional being could experience only part of a three dimensional object. A wormhole could connect extremely long distances such as a billion light-years or more, short distances such as a few meters, different universes, or different points in time. Wormholes are consistent with the general theory of relativity by Einstein, but whether wormholes actually exist remains a mystery. Now let’s look at the types of wormholes… The first type of wormhole solution discovered was the Schwarzschild wormhole, but it was found that it would collapse too quickly for anything to cross from one end to the other. These wormholes are non-traversable. It would be possible to keep it open only if exotic matter with negative energy density could be used to stabilize them.'' 

\noindent\textbf{Frame Captions:} ``a computer screen with a drawing of a large object types of worms

\noindent a red and yellow light streaks through a black background

\noindent a purple background with the words `scr' and a purple background with the words `sc'

\noindent a black and white image with the words `scrzid wombs' or einstein - riddle

\noindent a black background with the words `scr - røs '

\noindent a black and white circular image

\noindent a black and white image of a circular object

\noindent a circular image of clouds in the sky

\noindent a black hole in the sky with a dark background

\noindent a green vortex with a black hole in the center

\noindent a blue nebula with a white center and a black background

\noindent a blue nebula with a white center and a black center

\noindent a bright blue nebula with a bright star in the middle

\noindent a planetary nebula in the sky

\noindent a very blue and white cloud with a bright light

\noindent a bright blue sun shines through the clouds''

Given these inputs, the summary generated by GPT-3.5-Turbo is as follows. We observe that the noisy frame captions were refined into a correctly punctuated paragraph that supports the video transcript to describe what is likely happening in the video.

\begin{table}[!t]
\scriptsize
\begin{center}
\begin{tabular}{|*{5}{c|}}
\hline
Prompt & Transcript & Chapter Title & Frame Captions & Video Title\\
\hline
p1 &  1 & 0 &  0 & 0 \\
\hline
p2 & 0 & 1 & 0 & 0 \\
\hline
p3 & 0 & 0 & 1 & 0 \\
\hline
p4 &  0 & 0 & 0 & 1 \\
\hline
p5 & 0 & 0 & 1 & 1 \\
\hline
p6 & 0 & 1 & 0 & 1 \\
\hline
p7 &  0 & 1 &  1 & 0 \\
\hline
p8 & 1 & 0 & 0 & 1 \\
\hline
p9 & 1 & 0 & 1 & 0 \\
\hline
p10 & 1 & 1 & 0 & 0 \\
\hline
p11 & 0 &  1 & 1 & 1 \\
\hline
p12 & 1 & 0 & 1 & 1 \\
\hline
p13 & 1 & 1 & 0 & 1 \\
\hline
p14 & 1 & 1 & 1 & 0 \\
\hline
p15 & 1 & 1 & 1 & 1 \\
\hline
p16 & 1 & 1 & 0 & 0 \\
\hline
p17 & 1 & 1 & 0 & 1 \\
\hline
p18 & 1 & 1 & 0 & 0 \\
\hline
p19 & 1 & 1 & 0 & 1 \\
\hline
\end{tabular}
\caption{Prompt definitions based on inclusion or exclusion of specific signals in the prompt. For example, p5 utilizes only `Frame Captions' and `Video Title' in its prompt generation.}
\label{prompt_combiantions}
\end{center}
\end{table}

\begin{table}[!t]
\scriptsize
\centering
\begin{tabular}{|l|c|c|c|c|c|c|c|}
\hline
Prompt&B&C&M&D1&D2&BS&RL\\ \hline\hline
p1&28.7&0.239&21.0&39.3&83.5&56.0&13.0\\ \hline
p2&22.4&\textbf{0.844}&31.9&38.9&80.0&58.4&22.0\\ \hline
p3&18.5&0.109&15.2&34.0&71.3&46.7&13.2\\ \hline
p4&17.0&0.364&23.3&40.0&79.8&56.9&15.8\\ \hline
p5&\textbf{30.8}&0.371&24.2&43.3&\textbf{86.6}&60.0&18.0\\ \hline
p6&30.2&0.566&27.2&\textbf{43.8}&86.0&59.9&16.3\\ \hline
p7&16.2&0.340&24.9&34.3&70.3&53.5&18.3\\ \hline
p8&19.1&0.332&23.8&38.7&81.8&57.9&15.1\\ \hline
p9&11.0&0.320&19.4&33.0&73.6&53.6&13.3\\ \hline
p10&22.0&0.576&29.3&40.1&84.8&59.7&19.1\\ \hline
p11&18.4&0.636&29.0&38.6&81.5&58.3&\textbf{26.1}\\ \hline
p12&7.9&0.176&19.8&31.3&70.9&53.8&12.1\\ \hline
p13&21.2&0.615&28.2&39.6&83.0&\textbf{60.7}&18.6\\ \hline
p14&9.5&0.594&26.9&30.1&69.6&56.1&19.6\\ \hline
p15&13.8&0.284&22.9&34.1&75.1&54.8&16.4\\ \hline
p16&17.0&0.255&\textbf{45.8}&40.7&80.1&55.9&17.7\\ \hline
p17&19.3&0.660&26.6&40.3&80.0&56.6&18.6\\ \hline
p18&23.5&0.390&19.8&34.0&78.9&51.8&15.0\\ \hline
p19&15.4&0.250&17.7&33.7&75.6&50.6&12.8\\ \hline
\end{tabular}
\caption{A comprehensive comparison of different prompt combinations for Alpaca. Here, B=BLEU-1, C=CIDEr, M=METEOR, D1=Distinct-1, D2=Distinct-2, BS=BERT-Score, RL=ROUGE-L.}
\label{prompt_results}
\end{table}

\begin{table}[!t]
\scriptsize
\centering
\begin{tabular}{|l|c|c|c|c|c|c|c|}
\hline
Prompt&B&C&M&D1&D2&BS&RL\\ \hline\hline
p1 & 0.042 & 0.457 & 0.261 & 0.340 & 0.794 & 0.574 & 0.186 \\
\hline
p2 & \textbf{0.075} & \textbf{1.053} & 0.359 & 0.357 & 0.806 & 0.629 & \textbf{0.246} \\
\hline
p3 & 0.000 & 0.065 & 0.145 & 0.356 & 0.766 & 0.482 & 0.128 \\
\hline
p4 & 0.025 & 0.337 & 0.232 & \textbf{0.386} & \textbf{0.814} & 0.559 & 0.165 \\
\hline
p5 & 0.013 & 0.156 & 0.170 & 0.366 & 0.779 & 0.506 & 0.140 \\
\hline
p6 & 0.071 & 0.920 & \textbf{0.379} & 0.345 & 0.792 & 0.629 & 0.242 \\
\hline
p7 & 0.039 & 0.393 & 0.221 & 0.335 & 0.756 & 0.527 & 0.169 \\
\hline
p8 & 0.046 & 0.568 & 0.293 & 0.331 & 0.783 & 0.589 & 0.201 \\
\hline
p9 & 0.039 & 0.438 & 0.274 & 0.329 & 0.785 & 0.577 & 0.188 \\
\hline
p10 & 0.055 & 0.708 & 0.322 & 0.318 & 0.781 & 0.608 & 0.217 \\
\hline
p11 & 0.028 & 0.329 & 0.241 & 0.336 & 0.755 & 0.530 & 0.157 \\
\hline
p12 & 0.033 & 0.441 & 0.262 & 0.351 & 0.798 & 0.577 & 0.186 \\
\hline
p13 & 0.065 & 0.825 & 0.324 & 0.346 & 0.797 & 0.611 & 0.227 \\
\hline
p14 & 0.055 & 0.707 & 0.311 & 0.333 & 0.795 & 0.598 & 0.212 \\
\hline
p15 & 0.056 & 0.680 & 0.325 & 0.322 & 0.784 & 0.608 & 0.224 \\
\hline
p16 & 0.068 & 0.749 & 0.327 & 0.329 & 0.788 & 0.605 & 0.228 \\
\hline
p17 & 0.075 & 0.838 & 0.338 & 0.343 & 0.798 & 0.618 & 0.235 \\
\hline
p18 & \textbf{0.072} & 0.797 & 0.347 & 0.329 & 0.790 & 0.616 & 0.234 \\
\hline
p19 & 0.064 & 0.841 & 0.375 & 0.335 & 0.793 & \textbf{0.633} & 0.232 \\
\hline
\end{tabular}
\caption{A comprehensive comparison of different prompt combinations for GPT-3.5-Turbo. B=BLEU-1, C=CIDEr, M=METEOR, D1=Distinct-1, D2=Distinct-2, BS=BERT-Score, RL=ROUGE-L.}
\label{prompt_results_GPT3.5_Turbo}
\end{table}

\noindent\textbf{Summary:} The video likely discusses the concept of wormholes, theoretical gateways that could enable near-instantaneous travel to distant places in the universe, including different multiverses. The discussion raises questions about the possibility and safety of human proximity to wormholes. Wormholes are described as speculative structures connecting disparate points in spacetime, based on Einstein's field equations and the theory of general relativity.

The existence of wormholes remains to be practically discovered, leading many scientists to propose that they might be projections of a fourth spatial dimension. These hypothetical structures could potentially connect vast distances, ranging from billions of light-years to mere meters, allowing for travel between different universes or points in time. However, whether wormholes truly exist remains a mystery.

The video then delves into different types of wormholes, introducing the Schwarzschild wormhole as the first discovered solution. However, it is revealed that these wormholes would collapse too quickly for anything to traverse from one end to the other, making them non-traversable. The possibility of keeping them open is suggested by using exotic matter with negative energy density to stabilize them.

Although the video transcript provides a comprehensive explanation of wormholes, the frame captions focus more on various visual images, such as a computer screen displaying a drawing of a large object, different types of worms, streaks of red and yellow light against a black background, and purple and black backgrounds with cryptic words. Additionally, circular images of clouds, a black hole with a dark background, a green vortex surrounding a black hole, and various nebulae are mentioned, including blue nebulae with white centers, black centers, and a bright star in the middle. The frames also include a planetary nebula, a blue and white cloud illuminated by a bright light, and a bright blue sun shining through clouds. These images likely accompany or supplement the video's discussion on wormholes and the mysteries of the universe.


\section{BLIP vs ClipCap} \label{blipvsclipcap}

\begin{figure}

    \centering
    \includegraphics[width=0.35\textwidth]{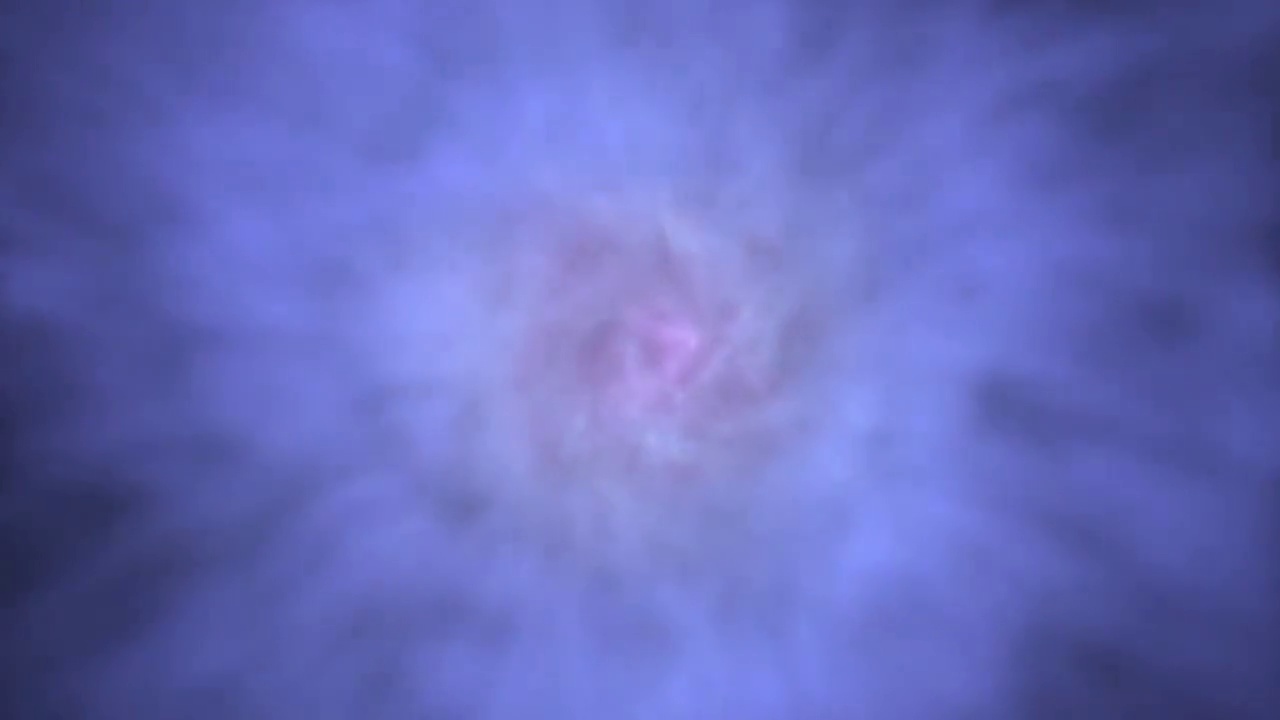}
    \caption{ClipCap Caption: travelling through a wormhole in deep space. BLIP Caption: a bright blue nebula with a bright star in the middle}
    \label{wormhole}
    \centering
    \includegraphics[width=0.35\textwidth]{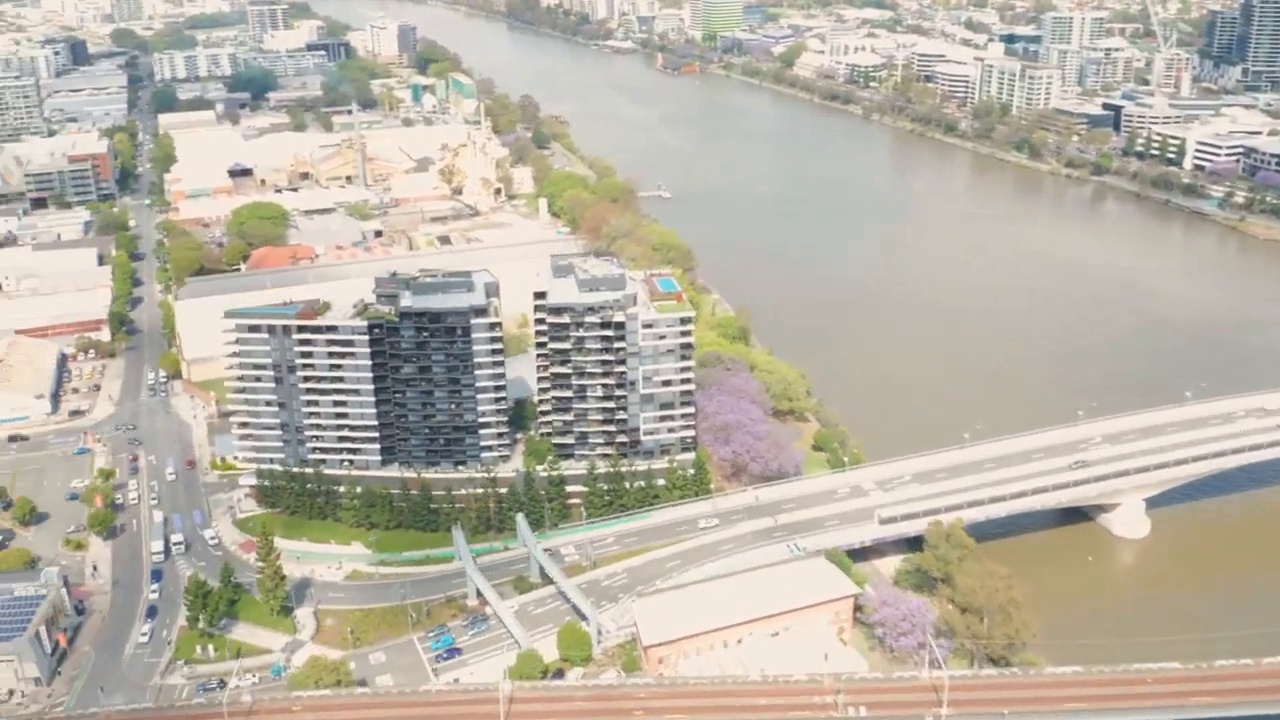}
    \caption{ ClipCap Caption: aerial view of the bridge over river. BLIP Caption: a bridge over a river with buildings in the background}
    \label{bridge}

\end{figure}

We noticed slight improvements ($\sim$0.07\%) with the BLIP image captioning model. This is because BLIP's captions offer more detailed information than ClipCap. For instance, in Fig. \ref{wormhole}, when describing an image of a wormhole, BLIP's caption includes details like the nebula, its color, a star, and its position, which are missing from the captions by ClipCap.

Similarly, in Fig. \ref{bridge}, for an image of a bridge in a cityscape, BLIP's caption mentions buildings that ClipCap overlooks. Thus, BLIP's ability to provide additional context improves our results.

\section{Prompt Engineering} \label{promt_engineering}
We explored a total of 19 combinations of inputs, considering the four input features: video transcript, frame captions, chapter title, and the video title. The complete set of input combinations is presented in Table \ref{prompt_combiantions}. We experiment with prompt engineering for Alpaca~\cite{taori2023stanford}
. 

First, we experiment with the following standard prompt template while providing all combinations of four features: transcript, frame captions, chapter title and video title. For example, considering chapter title and transcript as input, prompt is as follows. ``For the paragraph given below, bot is provided with the following attributes from a video: Chapter Title: [Chapter Title] and Transcript: [Transcript]. The question is then generated 
by utilizing the video data provided. The generated question should not contain any kind of pronouns. The generated Question's answer must be in the Chapter Title or Transcript. The generated question must have the highest attention. The generated Question must be as lengthy as possible. The generated question should be engaging, information-seeking, centered around some real world entity.'' This is the base prompt used in prompt templates p1 to p15.

Further, we also tried combinations of inputs with prompts where we provided positive (p18 and p19) or negative exemplar questions (p16 and p17). Prompt with positive exemplars augmented the above base prompt with this text: ``THE GENERATED QUESTIONS MUST BE LIKE: ``What is the history of the world’s largest hybrid tensegrity, Kurilpa Bridge?'', ``Does Syngonium Plant grow easily from properly taken cutting?'', ``How can you eat at a Khaja Ghar in
Nepal?'', ``How to wrap rice in banana leaf?'', ``What is the estimated net worth of Elizabeth Turner?'', ``Is it possible to travel through a wormhole, if they exist?'' Prompt with negative exemplars augmented the above base prompt with this text: ``DO NOT GENERATE QUESTIONS LIKE: ``What is the name of this place?'', ``Where do I need to take cuttings?'', ``What does she look like?'', ``What does dad do?'', ``What is her current net worth?'', ``What would happen if you take a trip to one of these wormholes?''

\begin{table}[!t]
\scriptsize
\centering
\begin{tabular}{|l|l|l|l|}
\hline
Model & Context Relevance & Engagement Index & Fluency \\ \hline
\citet{t5_neg_qs}&2.21&2.62&1.31\\ \hline
Alpaca-p2&2.52&2.82&3.44\\ \hline
GPT-3.5-Turbo-p2&3.69&3.72&\textbf{3.99}\\ \hline
$T5_{C}$(C)&2.93&3.50&3.69\\ \hline
$T5_{C}$(C, T)&3.13&3.26&3.35\\ \hline
$T5_{C}$(C, F)&3.01&3.24&3.35\\ \hline
$T5_{C}$(C, V)&3.38&3.34&3.37\\ \hline
$T5_{C}$(C, V, T)&3.12&3.30&3.45\\ \hline
$T5_{C}$(C, V, F)&3.44&3.49&3.53\\ \hline
$T5_{CC}$(C)&2.98&2.14&2.29\\ \hline
$T5_{CC}$(C, V)&3.30&3.35&3.36\\ \hline
$T5_{CC}$(C, T)&3.33&3.27&3.28\\ \hline
$T5_{CC}$(C, V, F, T)&3.68&3.71&3.61\\ \hline
$T5_{CC}$(C, V, S(F, T))&3.69&3.72&3.62\\ \hline
$B_{C}$ (C)&3.15&3.43&3.54\\ \hline
$B_{C}$ (C, V)&3.60&3.52&3.63\\ \hline
$B_{CC}$ (C)&3.39&3.44&3.46\\ \hline
$B_{CC}$ (C, V)&3.52&3.43&3.50\\ \hline
$B_{CC}$ (C, V, F, T)&3.69&3.74&3.62\\ \hline
$B_{CC}$(C, V, S(F, T))&3.75&3.84&3.72\\ \hline
$B_{CC}$(C, V, S(F, T), $E_C$) & \textbf{3.905} & \textbf{3.97} & \textbf{3.99}\\ \hline
\end{tabular}
\caption{Human Evaluation Results. Here, B=BART-large, T5=T5-base. C = Chapter Title, V = Video Title, F = Frame Caption, T = Transcript and S(F,T) = Summary of F and T, generated by GPT-3.5-Turbo.}\label{tab:human_evaluation_all}
\end{table}

\subsection{Prompt Design Considerations}

With an initial prompt, we observed that the model had a tendency to utilize demonstrative pronouns and pronouns in the generated questions, resulting in incomplete questions that lacked clarity in conveying the essence of the video. For instance, the model generated an ambiguous and incomplete question: ``\textit{What are some historical facts about this place?}'' To mitigate this issue, we incorporated specific directives in the prompt, instructing the model to avoid any pronouns or demonstrative pronouns. This led to the replacement of pronouns with appropriate nouns, resulting in improved questions. For example, the above question was transformed into ``\textit{What are some historical facts about Kurilpa?}''

To ensure that the generated questions included essential and significant keywords accurately representing the video chapters, we provided instructions in the prompt emphasizing attention and requesting the model to generate lengthy questions. By focusing on nouns, verbs, adjectives, and adverbs, which are crucial in conveying the main subject, action, or description, we aimed to include all necessary keywords. For instance, a question like ``\textit{What was the name of the temporary building during the 2018 Brisbane Olympics?}'' was enhanced to ``\textit{What was the name of the temporary building that housed the international broadcast centre during the 2018 Brisbane Olympics?}''
We included ``\textbf{Do}'' and ``\textbf{Do not}'' prompts to guide the model further, presenting examples of desired and undesired question formats. These examples helped shape the model's understanding of the expected question structure. 

\subsection{Prompt Evaluation} 
We evaluated the 19 prompts employing metrics such as BLEU-1, CIDEr, METEOR, BERT-Score, and ROUGE-L. Additionally, we experimented with two different frame caption-generating models: ClipCap and BLIP. We found BLIP to perform slightly better and hence have reported results using BLIP in the entire paper. We show results using each of the 19 prompts using Alpaca and GPT-3.5-Turbo in Tables~\ref{prompt_results} and~\ref{prompt_results_GPT3.5_Turbo} respectively. Different prompts lead to best results across different metrics. Overall we found prompt 2 to perform well across most metrics and hence show results using prompt 2 in the main results (Table~\ref{tab:ECISQGresults}).




\section{Divergence of T5}

Although $T5_{CC} (C, V, F, T)$ generally outperforms its counterpart, $T5_{CC} (C, V, S(F, T))$, across most automatic evaluation metrics, as depicted in Table \ref{tab:ECISQGresults}, it is important to note that $T5_{CC} (C, V, S(F, T))$ achieves higher Distinct 1 and 2 scores. This indicates that the model is capable of generating diverse questions when presented with well-structured and punctuated text (summary). While this diverse generation often leads to higher quality, as evidenced by human evaluation results in Tables \ref{human_evaluation} and \ref{tab:human_evaluation_all}, it doesn't perfectly align with the particular manually annotated gold questions, resulting in lower scores on similarity-based metrics such as BLEU-1, CIDEr, METEOR, BERT-Score, and ROUGE-L. Such observations are common in natural language generation use cases where multiple outputs can be correct, but ground truth typically has only one output. For example, several responses to an utterance in a conversation could be relevant in dialog modeling. Similarly, in question generation, several questions could be judged as relevant by human annotators, even though they may have little overlap with the single annotated question.

\section{Results of Various Domains}

To check for variation in metrics across various domains, we computed them across category-specific subsets of our test set. As mentioned in section \ref{Data Curation and Preprocessing}, of the 411 videos, the distribution of categories is as follows: Education: 121, Entertainment: 32, Howto \& Style: 90, News \& Politics: 8, People \& Blogs: 75, Science \& Technology: 65, Travel \& Events: 20. Table \ref{Results across various domains} shows results using the final model $B_{CC}(C, V, S(F, T), E_C)$. The results show that our model performs exceptionally well in the \textit{Entertainment} and \textit{Howto and Style} domains. We conducted our experiments using an A100 40 GB Nvidia GPU. It took approximately 2 hours to fine-tune the Bert-based classifier (Refer to Section \ref{Chapter Title Classifier}), and around 7-8 hours to develop the ECIS question generator (Refer to Section \ref{ECIS Questions Generator}).
\begin{table}

\scriptsize
\centering

\begin{tabular}{| l  |l  |l  |l  |l  |l  |l  |l |}
\hline
\textbf{Category} & \textbf{B} & \textbf{C} & \textbf{M} & \textbf{D1} & \textbf{D2} & \textbf{BS} & \textbf{RL} \\
\hline
Travel \& Events & 59.30 & 5.6120 & 67.74 & 48.85 & 89.55 & 80.03 & 65.85 \\
\hline
Howto \& Style & 77.15 & 7.5190 & 86.00 & 48.73 & 88.58 & 92.25 & 82.34 \\
\hline
Science \& Technology & 47.63 & 5.6073 & 65.33 & 45.04 & 86.35 & 83.75 & 64.80 \\
\hline
Entertainment & 89.17 & 9.1480 & 93.74 & 45.52 & 87.76 & 96.02 & 92.54 \\
\hline
People \& Blogs & 64.42 & 6.8570 & 76.83 & 48.92 & 87.95 & 87.57 & 72.86 \\
\hline
Education & 66.17 & 6.9368 & 79.13 & 45.10 & 86.34 & 88.29 & 75.95 \\
\hline

\end{tabular}
\caption{Results across various domains}\label{Results across various domains}

\end{table}

\section{Detailed Human Evaluation Protocol and Results}
To assess the quality of our generated questions, we conduct a human evaluation involving an expert with proficiency in English and a background in Computer Science and Engineering. Please note that this annotator is different from the one who initially annotated the dataset, ensuring the prevention of any potential biases. The evaluator was presented with a set of 100 generated question samples from each of our models, accompanied by their respective video and chapter information. We instruct the evaluator to rate each question on a scale of 0 to 4, assigning 0 to the poorest results and 4 to the best results, considering the following three criteria:
\noindent (1) Context Relevance: This criterion examines the extent to which the generated questions align with the content of the corresponding video chapter. It evaluates whether the questions are pertinent and directly related to the video's context.
\noindent (2) Engagement Index: This criterion gauges the level of engagement and interest elicited by the generated questions. The evaluator assessed whether the questions were captivating enough to capture the user's attention and foster a desire to explore the video further.
\noindent (3) Fluency: This criterion focuses on the grammatical correctness and coherence of the generated questions. The evaluator assessed the questions' linguistic fluency, ensuring they adhered to grammatical rules and were sound. By employing an expert evaluator and utilizing these three evaluation criteria, we obtain a comprehensive assessment of the quality of our generated questions.

Table~\ref{tab:human_evaluation_all} shows human evaluation results for several methods. In addition, to better understand subjectivity, we hired a new external annotator with a good grasp of the English language and an MBA degree to evaluate the same 100 samples of models $B_{CC}$(C, V, S(F, T), $E_C$) and $B_{CC}$(C, V, S(F, T), $E_R$). The kappa score for Context Relevance in both the models is above 90\%, indicating high agreement, while Engagement Index and Fluency lie in the respectable 60-70\% agreement scale. Thus, the average human evaluation kappa score for both models is 79\%, signifying a high degree of agreement. Table~\ref{tab:human-evaluation-kappa} shows detailed kappa scores.

\section{Detailed Related Work}
\label{app:relatedWork}

\subsection{Text-based Question Generation}
Text-based Question generation~\cite{pan2019recent,zhang2021review} has been studied from the following three main perspectives. (1) Input context text which could be at document level~\cite{pan2020semantic,yang2017semi,tuan2020capturing}, paragraph level~\cite{du2018harvesting,zhang2020dual}, sentence level~\cite{ali2010automatic} or keyword level~\cite{pan2020learning}. (2) Target answer: Methods could be agnostic of answer~\cite{chen2018learningq} or answer-aware. Answer aware methods could depend on an answer span~\cite{rajpurkar2016squad} highlighted in input context text or could be an abstract answer~\cite{bajaj2016ms}. (3) Generated question could be standalone questions, sequential questions (like a dialog)~\cite{reddy2019coqa} or multiple choice questions~\cite{gupta2022newskvqa}. 
Various QG studies have leveraged template/rule based methods~\cite{mazidi2014linguistic,lindberg2013generating,heilman2010good,duan2017question}, traditional sequence-to-sequence learning models like BiLSTMs~\cite{du2017learning,song2018leveraging} and Transformers~\cite{wang2020neural,chai2020learning,scialom2019self,kumar2019cross}, pre-trained Seq2Seq models~\cite{dong2019unified,xiao2021ernie,cheng2021guiding}, graph-based models~\cite{chen2023toward,su2020multi,liu2019learning,chen2020reinforcement,pan2020semantic}, and generative models like VAEs~\cite{wang2019answer,lee2020generating} and GANs~\cite{bao2018question,hosking2019evaluating,rao2019answer,yao2018teaching}.
Further, text-based QG has also been studied from a multi-hop perspective~\cite{su2020multi} where generated questions have multiple clauses and hence are more complex. In this work, we propose a novel video QG setting where we generate standalone questions in an answer-agnostic manner using Transformer-based models.

\subsection{Visual Question Generation}

\citet{mostafazadeh2016generating} introduced visual QG~\cite{patil2020visual} to generate questions from an image. Visual questions can be categorized in three groups: (1) visually grounded questions~\cite{antol2015vqa,krishna2017visual}, i.e., questions that can be answered based on information present in the image itself. (2) Commonsense-based questions~\cite{wang2017fvqa,wang2017explicit}, i.e.,  questions that can be answered using a combination of external commonsense knowledge source along with the grounded information in the image. (3) World knowledge-based questions~\cite{shah2019kvqa,penamakuri2023answer}, i.e., questions that can be answered using a combination of external factual knowledge base along with the grounded information in the image. Approaches used for visual QG include encoder-decoder models~\cite{mostafazadeh2016generating,zhang2017automatic}, compositional approaches~\cite{liu2018ivqa,patro2018multimodal,zhang2017automatic}, generative models~\cite{jain2017creativity}, reinforcement learning approaches~\cite{yang2018visual,fan2018reinforcement}, and bilinear pooling models~\cite{fukui2016multimodal,ben2017mutan,li2018visual}. Visual QG has also been studied in domain-specific ways~\cite{mehta2024circuitvqa}.


\subsection{Video Question Generation}

In video question generation~\cite{yang2021just,wang2020video,su2021end,guo2020multi}, the goal is to generate meaningful questions about a video optionally targeting an answer. These studies either generate questions only from transcripts~\cite{krishna2015automatic,huang2014tedquiz,priya2022automatic} or generate questions about common objects and attributes present in the video~\cite{yang2021just,wang2020video,guo2020multi,su2021end,gupta2022newskvqa,wang2021mirtt,lopez2020transformer}. Unlike these studies, we propose a novel problem of generating ECIS questions from videos. Further, previous studies rely solely on a single frame (visual information) to formulate the questions. Thus, they do not consider the spatiotemporal relationship among video frames. Unlike previous approaches, our QG model is not constrained to a specific domain. By leveraging multimodal information from both textual and visual sources, our model offers a more comprehensive and versatile approach to generate questions. 
\begin{table*}
\scriptsize
\centering

\begin{tabular}{| l  |l  |l  |l  |l |}
\hline
\textbf{Model} & \textbf{Context Relevance} & \textbf{Engagement Index} & \textbf{Fluency} & \textbf{Average} \\
\hline
\textbf{$B_{CC}$(C, V, S(F, T), $E_R$)} & 0.912 & 0.686 & 0.795 & 0.798 \\
\hline
\textbf{$B_{CC}$(C, V, S(F, T), $E_C$)} & 0.926 & 0.793 & 0.663 & 0.794 \\
\hline
\end{tabular}
\caption{Kappa scores for human evaluation} \label{tab:human-evaluation-kappa}
\end{table*}

\section{Qualitative Examples and Error Analysis}
\label{app:qualitative}
\noindent\textbf{Case Studies}: 
Table~\ref{tab:good_examples} showcases examples of questions generated by our top-performing method. While we regret the inability to include summaries of frame captions and transcripts due to space constraints, the table demonstrates the model's adeptness at integrating information from various sources—including video titles, chapter titles, and visual data—to produce relevant, engaging, and fluent questions that capture viewer interest.
In one case (not shown in the table), a frame caption read ``\textit{almonds in a bowl with the words -- almonds are a good source of magnesium}'', and a line from the summary was ``\textit{Lastly, there is a close-up of nuts, including almonds. Additionally, there is a doctor holding a model of a heart and red blood cells in the blood of a person, highlighting the importance of heart health.}'' The generated question, \textit{``What are the benefits of eating almonds daily?}'' effectively captures the video's intended message.

\noindent\textbf{Error Analysis}
To understand the kinds of mistakes our model does when generating questions, we manually categorized 100 generated questions based on the kind of errors, if any. We found four kinds of errors: (1) HA: Questions with Hallucinations (irrelevant/unrelated key words) (2) GR: grammatically incorrect questions (3) MK: Ambiguous questions or questions missing imperative keywords to make it an ECIS question (4) CX: 
Contextually incongruous questions which are not exactly related to the topic discussed in the video chapter. Table~\ref{bad_examples} shows a few of such examples. Amongst the overall errors, we found the following error type distribution: HA (38\%), GR (11\%), MK (27\%) and CX (24\%).

\noindent\textbf{LLAMA3-8B:}
We used the following two prompts for obtaining the zero-shot and fine-tuned results of LLAMA3-8B \cite{llama3modelcard}: 
(i) Prompt \textbf{P1} -``Below is an instruction that describes a task, paired with further context. Write a response that appropriately completes the request.
\#\#\# Instruction: Generate an Entity-centric Information seeking question from the Chapter\_Title, Transcript and Image\_Captions
\#\#\# Chapter\_Title:{}
\#\#\# Transcript:{}
\#\#\# Image\_Caption:{}
\#\#\# Response:{}''.
(ii) Prompt \textbf{P2} -``You are an expert english dataset annotator. Annotate a question such as ``What are the two main types of wormholes?'', that appropriately completes the following request.
\#\#\# Instruction: Generate a question from the Chapter\_Title, Transcript and Image\_Captions
\#\#\# Chapter\_Title:{}
\#\#\# Transcript:{}
\#\#\# Image\_Caption:{}
\#\#\# Response Question:{}''.

In Block A of Table \ref{tab:ECISQGresults}, Llama3-8B P1 and Llama3-8B P2 represent the zero-shot results obtained using Llama3-8B with prompts P1 and P2, respectively. It is important to note that the prompts provided to the model significantly influence the final results, as evidenced by the outcomes in Block A. Llama3-8B P2 in Block B of Table 2 shows the results obtained after fine-tuning Llama3-8B on our dataset using prompt P2, 3000 tokens and 50 epochs.

By providing the Transcript and Image Captions as input, we aimed to ensure the model was aware of information in both audio and video modalities. However, we observed that the model hallucinates or generates multiple ECIS/non-ECIS questions, leading to low similarity scores in the automatic evaluation metrics as shown in Table \ref{tab:ECISQGresults}, Block B. We highlight a few of such erroneous cases in Table \ref{tab:LLAMA3_BAD_examples_APPENDIX}.


In the first example (\textit{HJWPtFUsCiU}), LLAMA3-8B hallucinated and generated a JavaScript syntax instead of a question. In the second example (\textit{HYdBNdrDiX4}), the model generated multiple questions, all of which were ECIS in nature. However, in the third example (\textit{B4SSKjotkNc}), the model hallucinated and generated a url instead of a question. Additionally, in the example (\textit{YqAYmE8eoeQ}), although the model was able to generate an ECIS question, it included an extraneous suffix, ``\textit{Explain with examples},'' which does not conform to the structure of the annotated gold labels. Lastly, in the example (\textit{JwxIrvsb5RY}), the model generated multiple questions, some of which were non-ECIS in nature.

\begin{table*}[!t]
\scriptsize
\centering
\begin{tabular}{|p{0.65in}|p{1.45in}|p{1in}|p{1.5in}|p{1.25in}|}
\hline
VideoID&Video title&Chapter title& Gold question&Generate Question\\ \hline
\hline
Ky9lOmpvQ5Q & New Asus Vivobook S14 S433 Unboxing \& Review! | Best Laptop for Students \& Office Work 2020 & Touchpad \& Fingerprint 
& Does Asus Vivobook S14 S433 have a touchpad and fingerprint? & Does Asus Vivobook S14 S433 have a touchpad and fingerprint?\\ \hline

EgC0dlRUPMU & Nikon Z8 vs Nikon Z9 & Nikon Z8 vs Nikon Z9 
& How is the ``Body Type'', ``Sensor'', ``Processor'' of Nikon Z8 and Nikon Z9? &Which camera has better specs for photography Nikon Z8 or Nikon Z9?  
\\ \hline


2noPUBTmpK8 &Samsung Galaxy Z Fold5 vs Samsung Galaxy S23 Ultra vs iPhone 14 Pro Max & Display 
& What is the ``Display Type'', ``Display Size'' of Samsung Galaxy Z Fold5, Samsung Galaxy S23 Ultra and iPhone 14 Pro Max?  & How does the display of Samsung Galaxy Z Fold5 look? 
\\ \hline


unvzTHIXrNw & Simple, but delicious! A great New Year's snack that will amaze your guests & Onion 1 pc. 
& How to chop onion for tartlets? & How to chop onions? 
\\ \hline

jGxOAWRkkgU & How To Scale Your Consulting Business & 2 Ways to Grow
&  What are the 2 ways to grow your consulting business? & What are the 2 ways to grow your consulting business? \\ \hline

94UAKAPHDS0 & Clio Kill Cover NEW MATTE FOUNDATION + shade comparison | all shades swatch / how to / mask test & How to apply 
& How to evenly apply “Clio Kill Cover New Matte Foundation”? & How to apply Clio Kill Cover?
\\ \hline

pSRGErzzIo4 & How to Play Poker for Beginners PokerStars Learn & Hand rankings in poker 
& What is Hand rankings in poker? &What is Hand rankings in poker? \\ \hline


guU0hA0\_ST0 & Diddly Pay Pro Review - A Bryan Winters Product SURPRISED Where My Review Went [A MUST SEE!] & SURPRISE: AFTER I PURCHASED
& Is Clickbank not a secure website? & How can you refund money on Clickbank? \\ \hline

\end{tabular}
\caption{Example questions generated by our ``ECIS questions generator'' $B_{CC}$(C, V, S(F, T), $E_C$).} 
\label{tab:good_examples}
\end{table*}

\begin{table*}[!t]
\scriptsize
\centering
\begin{tabular}{|p{0.6in}|p{1.8in}|p{1in}|p{1.5in}|p{1.15in}|}
\hline
VideoID&Video title&Chapter title& Gold question&Generate Question\\ \hline
\hline
5T1MnBGX1hE & Resolve Issues Printing PDFs from Adobe Reader in Windows | HP Printers | @HPSupport & Change default program
& How to ‘change default program’? & How to change default program in Windows?\\ \hline

5VMLXnIx6lA & Walking tour of Heidelberg on a snowy day |Top sights of city \& castle | 
& Hauptstraße
& Which is the oldest and the longest path in Europe? & What is “Hauptstraße of Heidelberg” famous for?\\ \hline

6pY0p-nTmwo & WHY You SHOULD BUY (Not Lease) A TESLA MODEL 3 | 2021 Tesla Review & Over the Air Software Updates
& Is Tesla's software similar to IPhone? &Is Tesla's software similar to IPhone?\\ \hline

NZSEQz516g0 & What Beyblades sound like with a Maono AU 93 Microphone
& Why I need a microphone
& Why do you need microphones to listen to Beyblades? & Why should you invest in a microphone?\\ \hline

eLLAza87fzw & Here are 4 Courses Of Action To Heal Sacral Chakra & Blocked or Out of balance Sacral Chakra
& What does Blocked or Out of balance Sacral Chakra mean? & When can you feel drained or out of balance Sacral Chakra?\\ \hline

B4SSKjotkNc & What content should you include in your online membership?
& Extras
& How can a podcast be useful for making a long form version of a content? & What are the benefits online membership provides to build an online community?\\ \hline

th4PrDsh660 &Eng sub \begin{CJK}{UTF8}{min}
ボノボの「ホカホカ」はエロいのか？メス同士の性的？同性愛的？社会的行動？
\end{CJK} & ``Reconciliation'' in genito-genital rubbing\begin{CJK}{UTF8}{min} 喧嘩後の仲直りで「ホカホカ」\end{CJK}
& Why do female bonobos engage in ‘GG-rubbing for reconciliation’ after aggressive behaviors? & Why do female bonobos engage in ‘GG-rubbing for reconciliation’ after aggressive behaviors?\\ \hline 

eI67IMNqySo & 5 Profitable Video You Can Make With Your Phone
& Sales Funnel Video
& How can you make a 'Funnel Video with your phone'? & How can you make a 'Funnel Video with your phone'?\\ \hline

kivynV-TRYQ & Nespresso Creatista Uno Review | Nespresso Machine Reviews | A2B Productions & Price Dimensions and Specifications
& How is the Nespresso Creatista Machine? & How is the Nespresso Creatista Machine?\\ \hline

bbvoXbVY8VA & Celebrating Our Champion UFS Women 
& Sabeehah Vawda
& What are the thoughts of Sabeehah Vawda on the celebration of 'Our Champion UFS Women'? &What are the thoughts of Sabeehah Vawda on the celebration of 'Our Champion UFS Women'?\\ \hline

pm0s0NLQzH0 & Taking Care of Your Bottom after Baby & Dermoplast
& When can you use Dermoplast? & When can you use Dermoplast?\\ \hline

HJWPtFUsCiU & manifesting with the moon phases // moon manifestation
& new moon
& What does a 'New Moon' phase mean spritually? & What does a 'New Moon' phase mean spritually?\\ \hline

rwTKLb0de2w & Sony FX3 Setup for Hybrid Shooting - QUICKLY Change to Photo Mode! & Different Set for Still/Movie
& How to set up for Different Set for Still Movie in Sony FX3? & How to set up for Different Set for Still Movie in Sony FX3?\\ \hline

\end{tabular}
\caption{More example questions generated by our ``ECIS questions generator'' $B_{CC}$(C, V, S(F, T), $E_C$).} 
\label{tab:good_examples_APPENDIX}
\end{table*}

\begin{table*}[!t]
\scriptsize
\centering
\begin{tabular}
{|p{0.65in}|p{1.45in}|p{1in}|p{1.5in}|p{1.25in}|p{0.20in}|}
\hline
VideoID&Video title&Chapter title& Gold question&Generated Question & Error\\ \hline
\hline

hCWmRpK87J4 & How to Customize Your LinkedIn URL (Updated Nov 2021)  & Example of a LinkedIn URL 
& How to customize your LinkedIn profile URL? & What is the example of a LinkedIn URL? & CX
\\ \hline

\_ZNabbgpXgI & How to Get Rid of Oily Skin | 6 Tips to Cure Oily Skin | Tiege Hanley  & Eat Foods with Vitamin A 
& How does `food containing vitamin A' help oily skin? & How to eat foods with Vitamin A? & CX
\\ \hline

wfdcNfY0gxQ & Webcourses @UCF Student Tour  & Quizzes 
& How can you navigate to a quiz on UCF? & How to quizzes on UCF? & GR
\\ \hline

 

izOsczF16dA & 4 Phrygian mode chord progressions [Modal harmony \& songwriting tips]  & Chromatic mediants 
& What are Chromatic mediants in Phyrgian chord progression? & Are chromatic mediants 4 Phrygian mode chord progressions? & GR
\\ \hline

QkOS1OoI1Qg & SONY Z-V1 Mark II vs SONY Z-V1  & Camera Comparison 
& How is the ``Focus'', ``Shutter speed'' of SONY Z-V1 Mark II and SONY Z-V1? & Is there a microphone port in Sony Vaio Laptop? & HA
\\ \hline



hYyDBgj9yzY & CHARLIE CHAPLIN - The Champion (1915 HD) | Best Charlie Chaplin Comedy Videos | Silent Movie  & Edna the trainer's daughter. 
& Why did Edna, the trainer's daughter hugged Charlie Chaplin? & What is the author of the book CHARLIE CHAPLIN - The Champion (1915 HD)? & HA
\\ \hline

pcpxCOo\_8js & Music reactive LED | Sound sensor - Arduino Connections \& Coding  & Adjusting sensitivity of Sound sensor 
& How to adjust the sensitivity of a Sound sensor in Music Reactive LED? & How to adjust the sensitivity of a Sound sensor? & MK
\\ \hline

ShcCofKV6m4 & MFCL3770CDW – Connect to a mobile device  & Adding mobile device to the network 
& How to add a mobile device to the network of Brother Machine for printing? & How to add a mobile device to the network? & MK
\\ \hline

\end{tabular}
\caption{Examples of erroneous questions generated by our ``ECIS questions generator'' $B_{CC}$(C, V, S(F, T), $E_C$). Here, HA = Hallucination, GR = Grammatical Inaccuracy, MK = Missing Keyword and CX = Context Incongruity} 
\label{bad_examples}
\end{table*}

\begin{table*}[!t]
\scriptsize
\centering
\begin{tabular}{|p{0.6in}|p{0.8in}|p{1in}|p{1.2in}|p{1.25in}|p{1in}|}
\hline
VideoID&Video title&Chapter title& Gold question&Llama3-8B P2&$B_{CC}$(C, V, S(F, T), $E_C$)\\ \hline
\hline
HJWPtFUsCiU
 &manifesting with the moon phases // moon manifestation
 & new moon

&What does a 'New Moon' phase mean spritually?& 0) \{ return d.isReact; \} function hasSub subscription(c) \{ var d = c.subscription; if (d === und&What does a 'New Moon' phase mean spritually?\\ \hline

HYdBNdrDiX4
 & Experiment Builder Tutorial 10 - Screen Building and Referencing

& Screen Builder of a Display Screen Action Node

& How to add Screen Builder of a Display Screen Action Node in MindVAn? & How do we add content to a Display Screen Action Node? How to make reference to information in the Data Source to change Experimental characteristics across Trials? How to change the location of a Display Screen Action Node? &How to add Screen Builder of a Display Screen Action Node in MindVAn? \\ \hline

B4SSKjotkNc &What content should you include in your online membership?
 & A different format

& How to offer unique content by using different formats? &www.youtube.com/watch?v=H5 K8Y5WY8G8\&list=PLdQ8U7 zYBdS1t3gIaPbL7BwG3v7s & How to offer unique content by using different formats?\\ \hline

iruD9KTNnNc
& STRAIGHT LINE Method of Depreciation in 3 Steps!
& Example
& What is the example of ``Straight Line Depreciation''? & How to apply Straight-line depreciation method?  Explain with examples& What is the example of ``Straight Line Depreciation''?\\ \hline

JwxIrvsb5RY
& Jesus, the Most Extraordinary Man
& Public Ministry
& What did Jesus go through? & What was the ministry of Jesus? What did He do? In what ways did He minister to the people? What can we learn from Him?& How is the public ministry of Jesus?\\ \hline

\end{tabular}
\caption{A few example questions generated by the fine-tuned LLAMA3-8B P2 \cite{llama3modelcard}} 
\label{tab:LLAMA3_BAD_examples_APPENDIX}
\end{table*}


\begin{table*}[!t]
\scriptsize
\centering
\begin{tabular}
{|p{0.60in}|p{1.5in}|p{1in}|p{1.5in}|p{1.25in}|p{0.20in}|}
\hline
VideoID&Video title&Chapter title& Gold question&Generated Question & Error\\ \hline
\hline
JATZvVIuSGs & How to Create Form to Collect Customer Data in PageFly \#1 Shopify Page Builder  &  Configuration in Customer Form field elements 
& How to 'Configure the Customer Form field elements' in PageFly? & How to configure in Customer Form field elements? & Cx
\\ \hline

n5GvKSM92Ic & 6 What Am I Riddles with Answers to Test Your Brain  & I cannot be burned in fire or drowned in water. 
& What does 'I cannot be burned in fire or drowned in water' mean? & Can you be burned in fire or drowned in water?  & Cx
\\ \hline

n5GvKSM92Ic & 6 What Am I Riddles with Answers to Test Your Brain  & I am a ship that can be made to ride the greatest waves. I am not built by objects but built by minds. 
& What does 'I am a ship that can be made to ride the greatest waves. I am not built by objects but built by minds' mean? & How is a ship made to ride the greatest waves? & Cx
\\ \hline

EgC0dlRUPMU & Nikon Z8 vs Nikon Z9  & Nikon Z9 
& How is the ``Display'' of Nikon Z8 and Nikon Z9? & Which camera has better dimensions, Nikon Z8 or Nikon Z9? & Cx
\\ \hline

hYyDBgj9yzY & CHARLIE CHAPLIN - The Champion (1915 HD) | Best Charlie Chaplin Comedy Videos | Silent Movie  & Completely broke.Meditating on the ingratitude of humanity. 
& Why did the Dog didn't eat the sausage when Charlie Chaplin offered one? & How can you meditate on the ingratitude of humanity? & Cx
\\ \hline

6pY0p-nTmwo & WHY You SHOULD BUY (Not Lease) A TESLA MODEL 3 | 2021 Tesla Review  & Very Few Mechanical Issues to Worry About 
& What are the mechanical issues in Tesla that you should worry about? & What are the reviews for Tesla's Model 3? & Cx
\\ \hline

JwxIrvsb5RY & Jesus, the Most Extraordinary Man  & Public Ministry 
 & What did Jesus go through? & How is the public ministry of Jesus? & Cx
\\ \hline

eLLAza87fzw & Here are 4 Courses Of Action To Heal Sacral Chakra  & Blocked or Out of balance Sacral Chakra 
 
& What does Blocked or Out of balance Sacral Chakra mean? & When can you feel drained or out of balance Sacral Chakra? & Gr
\\ \hline

VyogSFRV\_k8 & Winnie the Pooh coin collection by royal mint  & Christopher Robin 50p 
& Where can you get Christopher Robin 50p? & What is the book 'Winnie the Pooh'? & Ha
\\ \hline

JOc2GhTCVjU & HOW THE CASHIER CALCULATES THE CHANGE QUICKLY (WITH EXAMPLES)  & How I count to myself (IRL) 
& How can you count cash money yourself? & How can you add your customer's money to yourself? & Ha
\\ \hline

ftFs6alnwYo & [Eng] \begin{CJK}{UTF8}{mj}호주 브리즈번 일상 브이로그 \end{CJK} | Brisbane vlog, \begin{CJK}{UTF8}{mj}  뉴스테드 카페 브런치, 보웬힐 킹스트리트 저녁데이트 \end{CJK} | \begin{CJK}{UTF8}{mj} 아이폰12프로 언박싱 \end{CJK}  & \begin{CJK}{UTF8}{mj} 브리즈번 애플스토어 \end{CJK} Brisbane Apple Store 
 
 
& How is the Brisbane Apple Store? & What are the features of the Asus Vivobook 6s? & Ha
\\ \hline

jbNau9\_Rmqk & Restless (\begin{CJK}{UTF8}{gbsn}心神不宁\end{CJK}) | S1E4 | One Word A Day  & Review of ``Intoxicated'' word
& How do you say `Obsessive Mindset Grandiose Thinking' in Chinese? & How does the book ``Intoxicated'' word Restless (\begin{CJK}{UTF8}{gbsn}心神不宁\end{CJK}) cause death? & Ha
\\ \hline

n5GvKSM92Ic & 6 What Am I Riddles with Answers to Test Your Brain  & I have roots nobody sees. I am taller than trees. Up, up I grow but I never grow 
& What does 'I have roots nobody sees. I am taller than trees. Up, up I grow but I never grow' mean? & What kind of roots does a tree grow higher than a man? & Ha
\\ \hline

QkOS1OoI1Qg & SONY Z-V1 Mark II vs SONY Z-V1  & SONY Z-V1 Mark II vs SONY Z-V1 
& How is the ``Body Type'',``Sensor'', ``Processor'' of SONY Z-V1 Mark || and SONY Z-V1? & Which camera has better specs for photography Sony Z-V1 Mark II or SONY Z-VR1? & Ha
\\ \hline

QkOS1OoI1Qg & SONY Z-V1 Mark II vs SONY Z-V1  & SONY Z-V1 Mark II 
& How is the ``Resolution'', ``ISO Sensitivity'' of SONY Z-V1 Mark || and SONY Z-V1? & Which camera has a better LCD display, SONY Z-V1, or SONY Laptop? & Ha
\\ \hline

QkOS1OoI1Qg & SONY Z-V1 Mark II vs SONY Z-V1  & SONY Z-V1 
&How is the ``Display'' of SONY Z-V1 Mark II and SONY Z-V1?  &Which camera has better dimensions, Sony Vaio Laptop or SONY Z-V1? & Ha
\\ \hline

QkOS1OoI1Qg & SONY Z-V1 Mark II vs SONY Z-V1  & \#TRENDING 
& How is the ``Weight'', ``Dimension'' and ``Battery Life'' of SONY Z-V1 Mark II and SONY Z-V1? & Which camera has better dimensions, Sony Vaio Laptop, or SONY Z-V1?  & Ha
\\ \hline

7elGaUPAhRQ & SONY FX30 vs SONY FX3  & \#TRENDING 
&How is the ``Weight'', ``Dimension'' and ``Battery Life'' of SONY FX30 and SONY FX3? & What happened in 1983 in Indian cricket history? & Ha
\\ \hline

hYyDBgj9yzY & CHARLIE CHAPLIN - The Champion (1915 HD) | Best Charlie Chaplin Comedy Videos | Silent Movie  & The big night. 
 & Why were so many people in the boxing ring along with Charlie Chaplin? & What is the history of `Chrome'? & Ha
\\ \hline

z0cB\_ui4xXY & Oppo Find X6 Pro Vs. Samsung Galaxy S23 Ultra Vs. Apple iPhone 14 Pro Max  & Platform 
& What is the ``Operating System'', ``Chipset'', ``Processor'' of Oppo Find X6 Pro, Samsung Galaxy S23 Ultra and Apple iPhone 14 Pro Max? & Which camera has max resolution Panasonic Vaio Laptop? & Ha, Gr
\\ \hline

Ow9taJw2gz0 & DIOR Capture Dreamskin Cushion 020 030  & What's inside the cushion? 
&What is inside the cushion of Dior Capture Shade 030? & What's inside the cushion? & Mk
\\ \hline

R1p2f\_afO74 & The Story Of Nick Leeson | Rogue Trader | SIMEX  & The Quintuple 8 account 
& How did Nick Leeson play with The Quintuple 8 account? & What is the Quintuple 8 account? & Mk
\\ \hline

hCWmRpK87J4 & How to Customize Your LinkedIn URL (Updated Nov 2021)  & How to use keywords 
&How can you customize your LinkedIn URL to be in the top results? &How can you use keywords? & Mk
\\ \hline

CY8IDagpUJA & iPhone XS Max Battery Replacement  & LCD Assembly Removal 
& How to remove the LCD Assembly of XS Max? & How to remove the LCD display? & Mk
\\ \hline

hYyDBgj9yzY & CHARLIE CHAPLIN - The Champion (1915 HD) | Best Charlie Chaplin Comedy Videos | Silent Movie  & Training for his first fight. 
& What Excercise did Charlie Chaplin do while training to be a champion? & How to train for his first fight? & Mk
\\ \hline

hYyDBgj9yzY & CHARLIE CHAPLIN - The Champion (1915 HD) | Best Charlie Chaplin Comedy Videos | Silent Movie  & Five big ones if you let yourself lose. 
& How did the big Moustache guy felt down? & What happens if you let yourself lose?& Mk
\\ \hline

\end{tabular}
\caption{More examples of erroneous questions generated by our ``ECIS questions generator'' $B_{CC}$(C, V, S(F, T), $E_C$). Here, HA = Hallucination, GR = Grammatical Inaccuracy, MK = Missing Keyword and Cx = Context Incongruity} 
\label{bad_examples_APPENDIX}
\end{table*} 
 



\noindent\textbf{Comparison with Qwen-VL and GPT-4o:}

We highlight a few samples below for a clear comparison of our best model with Qwen-VL and GPT-4o in Table~\ref{tab:comparisonQwenGPT4o}.

\begin{table*}[!t]
    \centering
    \scriptsize
    \begin{tabular}{|l|p{0.2\textwidth}|p{0.2\textwidth}|p{0.2\textwidth}|p{0.2\textwidth}|}
\hline
Video ID&Gold Question&Qwen-VL&GPT-4o&$B_{CC}(C, V, S(F, T)$\\
\hline
\hline
90pPhEqe0ck&How to create a ``House Wholesaling Checklist''?&What is the process of wholesaling a house, and how can you download our free checklist to get started?&Where can I download the free wholesaling checklist promoted by the man in a red shirt?&How to create a ``House Wholesaling Checklist''?\\
\hline
s7OQ2oNr6E4&How to make the sausage gravy for buttermilk substitute?&What is the recipe for making buttermilk biscuits and gravy from scratch?&How does the person prepare sausage gravy in the video?&How to make a sausage gravy for buttermilk substitute?\\
\hline
nRigYU13rig&How is Mandarin Oriental KL?&What is the name of the hotel where the video shows people taking selfies and getting hair done?&What activities are people engaging in at the Mandarin Oriental KL as shown in the video?&How is the Mandarin Oriental in KL?\\
\hline
    \end{tabular}
    \caption{Comparison of our best model with Qwen-VL and GPT-4o}
    \label{tab:comparisonQwenGPT4o}
\end{table*}

In the case of video ID: 90pPhEqe0ck, both Qwen-VL and GPT4o fail to generate a self-complete question that is sensible to a user without additional context. The phrases `our free checklist' in Qwen-VL's generation and `by the man in a red shirt?' in GPT4o's generation make them ambiguous.

In the case of video ID: s7OQ2oNr6E4, GPT4o's generation is not a self-complete question due to the ambiguous phrase `gravy in the video?.'

In the case of video ID: nRigYU13rig, both Qwen-VL and GPT4o fail to generate a self-complete question. Qwen-VL's generation includes the ambiguous phrase `where the video shows,' and GPT4o's generation includes `as shown in the video?.'

\section{Ablation for Input Representations} 
\label{app:ablation}
Table~\ref{tab:ablations} shows results for ECIS question generation using different combinations of input representations. We also show results using BART/T5 and using just cross-entropy loss vs. a combination of contrastive and cross-entropy loss. Introducing contrastive loss yields superior results compared to exclusively relying on cross-entropy. This trend is observed across all the models with different input combinations as well as for both BART and T5.
We also observe that models with the input features (C,V,T) and (C,V,S(F,T)) achieve the best results, with (C,V,T) performing exceptionally well in BLEU-1, CIDEr and ROUGE-L. Although (C,V,S(F,T)) is slightly inferior in automatic evaluation metrics, it generates higher-quality questions according to the human evaluation results as shown in Table~\ref{human_evaluation}.




Further, for our best model, $B_{CC}$(C, V, S(F, T)) as well as the equivalent T5 variant, we tried two variations by including video embeddings obtained using CLIP or ResNeXt as an additional input. Table~\ref{tab:ablation2} shows that using CLIP based video embeddings is better than using ResNeXt based ones for BART but ResNeXt embeddings are better for T5. Overall, $B_{CC}$(C, V, S(F, T)) with $E_C$ is the best.

\begin{table*}[htbp]
\begin{minipage}[b]{0.48\textwidth}
\centering
\scriptsize
\begin{tabular}{|l|c|c|c|c|c|c|c||c|c|c|c|c|c|c|}
\hline
\multirow{2}{*}{Model}&\multicolumn{7}{c||}{Cross-Entropy Loss}&\multicolumn{7}{c|}{Contrastive Loss + Cross-Entropy Loss}\\ \cline{2-15}
&B&C&M&D1&D2&BS&RL&B&C&M&D1&D2&BS&RL\\ \hline\hline
$T5$(C)&15.8&1.8&32.1&39&65.8&59.1&29.2&27.1&2.9&42.2&40.3&72.8&65&39.3\\ \hline
$T5$(C, V)&28.3&3&52.2&48.1&87.2&73.3&44.4&47.9&5.2&64.9&49.2&87.9&80.8&60.6\\ \hline
$T5$(C, F)&26.7&2.9&49.8&47.2&87.1&71.8&42.7&60.6&6.3&73.3&47.4&86.9&84.6&70.3\\ \hline
$T5$(C, T)&23.4&2.5&47.7&45.4&86.3&70.3&39.6&58.2&5.9&71.1&46.6&86.8&83.1&67.0\\ \hline
$T5$(C, V, F)&30.7&3.2&54.8&48.1&88.1&75.2&47.2&39.3&4.1&59.3&47.2&87.3&76.3&52.7\\ \hline
$T5$(C, V, T)&22.8&2.5&48.3&48.8&88.1&71.9&41.2&52.3&5.5&69.5&49.1&87.9&83.4&64.5\\ \hline
$T5$(C, F, T)&24.9&2.5&48.4&45.5&86.0&71.0&41.1&63&6.5&75.4&46.7&87.1&86.3&71.8\\ \hline
$T5$(C, V, F, T)&26.1&2.8&51.3&48.2&87.7&73.8&43.6&59.4&6.0&72.8&47.9&87.9&85.1&69.5\\ \hline
$T5$(C, V, S(F, T))&29.4&3.2&61.9&49.0&88.5&80.3&46.5&49.3&5.5&70.7&49.1&88.5&85.1&62.2\\ \hline \hline
$B$(C)&26.8&2.9&49.6&47.3&87.5&72&43.2&62.1&6.5&74.2&47.7&87.8&86&71.4\\ \hline
$B$(C, V)&31.7&3.4&57.4&48.8&88.3&77.6&49.7&58.1&6.1&74.4&48.1&87.3&86.1&70.3\\ \hline
$B$(C, F)&30.2&3.3&54.6&47.2&87.5&75.4&47.7&45.1&4.8&64.1&48.5&87.7&80.4&59.7\\ \hline
$B$(C, T)&22.9&2.4&48.1&45.7&87.1&70.6&40.6&68.5&7.1&79.3&46.8&87.4&88.4&77.0\\ \hline
$B$(C, V, F)&30.1&3.2&55.6&44.7&85.9&75.1&47.6&46.4&5.1&66.4&48.3&87.3&81.8&62.7\\ \hline
$B$(C, V, T)&30.8&3.3&57.1&48.9&88.6&77.2&48.9&\textbf{70.6}&\textbf{7.3}&81.4&47.2&87.7&89.4&\textbf{79.0}\\ \hline
$B$(C, F, T)&32.1&3.4&55.9&46.9&87.3&76.1&50.1&58.3&6.1&72.8&48.4&87.8&85.6&69.9\\ \hline
$B$(C, V, F, T)&29.7&3.2&54.4&47.7&86.8&75.3&48&54.4&5.7&71&48.6&87.5&84.4&66.9\\ \hline
$B$(C, V, S(F, T))&28.2&3.6&56.4&49.7&88.6&73.5&43.2&67.8&7.1&\textbf{83.5}&\textbf{50.4}&\textbf{88.9}&\textbf{91.3}&76.9\\ \hline
    \end{tabular}
    \caption{Ablation results of ECIS question generation. 
}
    \label{tab:ablations}
    \end{minipage}
    \hfill

\begin{minipage}[b]{0.45\textwidth}
\centering \scriptsize
\begin{tabular}{| l | l | l | l | l | l | l | l |}
\hline
\textbf{Model} & \textbf{B} & \textbf{C} & \textbf{M} & \textbf{D1} & \textbf{D2} & \textbf{BS} & \textbf{RL} \\
\hline
\citet{t5_neg_qs} & 35.15 & 3.9 & 60.51 & 47.81 & 87.95 & 79.58 & 51.84 \\
\hline
\citet{lopez2020transformer} & 24.30 & 2.7 & 52.24 & 47.21 & 88.65 & 73.56 & 41.86 \\
\hline
\citet{vqg_baseline_lmqg} & 27.67 & 3.3 & 54.82 & 47.76 & 88.25 & 75.86 & 46.27 \\
\hline
\citet{yang2021just} & 36.62 & 4.1 & 62.93 & 47.36 & 87.49 & 80.03 & 53.38 \\
\hline
\hline
$B_{CC}$(C, V, S(F, T), $E_C$)&\textbf{71.3}&\textbf{7.3}&\textbf{81.9}&47.2&87.6&\textbf{90.00}&\textbf{78.60}\\ \hline
\end{tabular}
\caption{Results of baselines after finetuning on our dataset} \label{tab:baseline-finetuning}
\end{minipage}%
\hfill
\begin{minipage}[b]{0.48\linewidth}
\centering
\scriptsize
\begin{tabular}{|l|c|c|c|c|c|c|c|}
\hline
Model&B&C&M&D1&D2&BS&RL\\ \hline\hline
$T5_{CC}$(C, V, S(F, T))&49.3 & 5.5 & 70.7&49.1&88.5&85.1&62.2\\ \hline
$T5_{CC}$(C, V, S(F, T), $E_C$)&43.3 & 4.9 & 64.5&49.4&88.1&81.3&60.0\\ \hline
$T5_{CC}$(C, V, S(F, T), $E_R$)&49.5 & 5.6 & 69.5&49.8&88.3&84.1&65.8\\ \hline
$B_{CC}$(C, V, S(F, T))&67.8&7.1&\textbf{83.5}&\textbf{50.4}&\textbf{88.9}&\textbf{91.3}&76.9\\ \hline
$B_{CC}$(C, V, S(F, T), $E_C$)&\textbf{71.3}&\textbf{7.3}&81.9&47.2&87.6&90.0&\textbf{78.6}\\ \hline
$B_{CC}$(C, V, S(F,T), $E_R$)&70.3&7.2&81.3&46.7&87.5&89.6&77.9\\ \hline
\end{tabular}
\caption{Impact of video embeddings. $E_C$ ($E_R$) = Embeddings obtained using CLIP (ResNeXt).}
\label{tab:ablation2}
\end{minipage}
\end{table*}

\section{Frequently Asked Questions (FAQs)}

\noindent$\bigast$ \textbf{What are the compute resource requirements for training/fine-tuning the system?}

$\Rightarrow$ We have done experiments on A100 40 GB Nvidia GPU. Fine-tuning the Bert-based chapter title classifier takes ~2 hours, while the training for ECIS question generator takes ~7-8 hours.

\noindent$\bigast$ \textbf{Are the annotators also the authors?}

$\Rightarrow$ No, we have used internal annotators. We also hired external annotators for both the chapter title classifier and question generation annotations.




\noindent$\bigast$ \textbf{What was the value of $\lambda$? (Refer Section \ref{Fine-tuning using combination of Contrastive and Cross-entropy Loss})}

$\Rightarrow$ We set $\lambda = 1$ to give the same weight to cross entropy as well as contrastive loss.

\noindent$\bigast$ \textbf{Does the proposed system compromise computational efficiency when compared to existing single-model solutions?}

$\Rightarrow$ Please note that our model is trained end-to-end and the loss computation part of our pipeline happens only at train time. 

At inference time, we agree there are several components that need to run (like BERT classifier, frame extraction, GPT3.5 summary generation, CLIP for video embedding, T5 for final question generation). However, it is important to note that question generation is typically an offline application, so the inference time of the entire pipeline might not be a significant concern, particularly if accuracy enhancements are substantial. One of our primary goals has been to develop a resilient question generation system.

\noindent$\bigast$ \textbf{ What is the source of the ``Gold Question'' mentioned in the paper? }

$\Rightarrow$ Two annotators manually annotated the Gold Questions in their late 20s. Both annotators have a Bachelor's in Computer Science and Engineering and a good grasp of English, with prior knowledge of similar tasks. The specific annotation guidelines are highlighted in Section~\ref{dataset}.

\noindent$\bigast$ \textbf{ What does the phrase ``self-complete'' mean? }

$\Rightarrow$ The term ``Self-complete'' refers to a phrase or question that can be comprehended independently without requiring additional context information. For instance, questions like ``What is a Wormhole?'' or ``What is Servant Leadership in Business?'' are self-complete as their meanings can be readily understood. On the contrary, a question like ``When was she born?'' is not self-complete since the subject being referred to lacks clarity without additional context information clarifying the individual's identity.

\noindent$\bigast$ \textbf{ Why were the UL questions dropped?}

$\Rightarrow$ UL stands for useless. We train the classifier over the dataset \textbf{without dropping} the UL class. This ensured that the classifier could identify all of the chapter types. But during the generation phase, if a chapter has been \textbf{classified as UL, then we discard it}. This is because the video chapter corresponding to such chapter title types does not hold informative content, as mentioned in Section \ref{dataset}. If there is no informative content, there is no way one can use it to generate an ECIS question. Hence, we discard these from the generation phase.

\noindent$\bigast$ \textbf{ Is there any over-fitting phenomenon? Is training done in two-stage or is it joint training for the classification and generation task?}

$\Rightarrow$ We incorporated dropout layers with a probability of 0.2 after every linear layer. Additionally, we utilized a specific optimization strategy by defining different weight decay values for different parameter groups in the AdamW optimizer, with a learning rate of 3e-5. These measures were implemented to mitigate the risk of over-fitting and promote generalization during training of the chapter title classifier.

It was a two-stage training; specifically, we first trained the classifier and then the generation model.



\noindent$\bigast$ \textbf{Why are SCQ questions retained?}

$\Rightarrow$ While SCQs are included in the dataset, our generation model isn't trained on them. When the Chapter title type is SCQ (as shown in Fig. \ref{model}), we directly output it without involving our BART/T5/Alpaca models. Since these are directly usable as ECIS questions, we do not need to pass them through Transformer-based generative models.

\end{document}